%% file: COLI_template.tex
\documentclass[final]{clv2025}

\jvol{vv}
\jnum{nn}
\jyear{2025}

\dochead{} 

\pageonefooter{Action editor: \{action editor name\}. Submission received: DD Month YYYY; revised version received: DD Month YYYY; accepted for publication: DD Month YYYY.}
\usepackage{float}
\usepackage{amsmath}
\usepackage{booktabs}
\usepackage{multirow} 
\usepackage{booktabs}
\usepackage{makecell}
\usepackage{subcaption}
\usepackage{soul}
\usepackage{times}
\usepackage{hyperref}
\usepackage{latexsym}
\usepackage{enumitem}
\usepackage[T1]{fontenc}
\usepackage{lineno} 

\newcommand{\cljournalquestion}[1]{\textcolor{brown}{}}
\newcommand{\cljournalanswer}[1]{\textcolor{black}{#1}}

\usepackage[utf8]{inputenc}
\usepackage{natbib}
\usepackage{microtype}
\usepackage{tikz}
\usepackage[normalem]{ulem}
\usepackage{inconsolata}
\usepackage{amsmath}
\usepackage{graphicx}
\usepackage{xcolor} 
\newcommand{\laks}[1]{}
\newcommand{\LL}[1]{\textcolor{black}{#1}}

\newcommand{\rudraorange}[1]{}
\newcommand{\rudrabluetext}[1]{\textcolor{black}{#1}}

\newcommand{\eat}[1]{}
\newcommand{\baselinei}{AEC}
\newcommand{\baselineii}{KPA}
\newcommand{\baselineiii}{KPA-GPT}
\newcommand{\ourapproachbest}{STIC*}
\newcommand{\ourapproachroberta}{STIC-R}
\newcommand{\ourapproachmst}{STIC-D}

\newcommand{\squishlist}{
 \begin{list}{$\bullet$}
  { \setlength{\itemsep}{0pt}
     \setlength{\parsep}{3pt}
     \setlength{\topsep}{3pt}
     \setlength{\partopsep}{0pt}
     \setlength{\leftmargin}{1.5em}
     \setlength{\labelwidth}{1em}
     \setlength{\labelsep}{0.5em} } }

\newcommand{\squishlisttwo}{
 \begin{list}{$\bullet$}
  { \setlength{\itemsep}{0pt}
    \setlength{\parsep}{0pt}
    \setlength{opsep}{0pt}
    \setlength{\partopsep}{0pt}
    \setlength{\leftmargin}{2em}
    \setlength{\labelwidth}{1.5em}
    \setlength{\labelsep}{0.5em} } }

\newcommand{\squishend}{
  \end{list}  }
\newsavebox{\myenumbox}
\newcommand{\reditem}{\stepcounter{enumi}\item[\color{red}\arabic{enumi}.]}
\newcommand{\greenitem}{\stepcounter{enumi}\item[\color{green}\arabic{enumi}.]}

\runningtitle{A Community-Based Approach for Stance Distribution and Argument Organization}
\runningauthor{Saha}
\begin{document}

\title{A Community-Based Approach for Stance Distribution and Argument Organization}

\author{Rudra Ranajee Saha\thanks{Corresponding authors}$^{*,1}$, Laks V. S. Lakshmanan$^{2}$, Raymond T. Ng$^{3}$}

\affilblock{
    \affil{University of British Columbia\\\quad \email{rrs99@cs.ubc.ca}}
    \affil{University of British Columbia\\\quad \email{laks@cs.ubc.ca}}
    \affil{University of British Columbia\\\quad \email{rng@cs.ubc.ca}}
}




\maketitle

\begin{abstract}
{\rm 
The proliferation of online debate platforms and social media has led to an unprecedented volume of argumentative content on controversial topics from multiple perspectives. While this wealth of perspectives offers opportunities for developing critical thinking and breaking filter bubbles \citep{pariser2011filter}, the sheer volume and complexity of arguments make it challenging for readers to synthesize and comprehend diverse viewpoints effectively. We present an unsupervised graph-based approach for community-based argument organization that helps users navigate and understand complex argumentative landscapes. Our system analyzes collections of topic-focused articles and constructs a rich \textit{interaction graph} by capturing multiple relationship types between arguments: topic similarity, semantic coherence, shared keywords, and common entities. We then employ community detection to identify argument communities that reveal homogeneous and heterogeneous viewpoint distributions. The detected communities are simplified through strategic graph operations to present users with digestible, yet comprehensive summaries of key argumentative patterns. Our approach requires no training data and can effectively process hundreds of articles while preserving nuanced relationships between arguments. Experimental results demonstrate our system's ability to identify meaningful argument communities and present them in an interpretable manner, facilitating users' understanding of complex socio-political debates. } 
\end{abstract}

\input{introduction}
\input{related_works}
\input{problem_formulation}
\input{methodology}

\input{experiment}

\input{conclusion}

\input{limitations}

\appendix
\input{appendix}

\bibliographystyle{compling}
\bibliography{custom}

\end{document}

%% file: introduction.tex
\section{Introduction}
The digital age has transformed how individuals consume information about controversial topics. Modern search engines and social media platforms, designed to maximize user engagement, often create "filter bubbles" by prioritizing content aligned with users' existing beliefs and preferences. These filter bubbles isolate users from diverse perspectives, posing a significant challenge to informed and constructive dialogue across stakeholders of diverse viewpoints, as well as policy decision-making by systematically limiting exposure to counterarguments \citep{ekstrom2022self, wolfowicz2023examining, ross2022echo}.
\par
Understanding opinions on controversial topics begins with identifying stance—the position an author takes toward a claim, typically classified as supporting, opposing, or remaining neutral. Stance detection, which has been extensively studied in computational linguistics, focuses on automatically identifying these positions in individual articles or social media posts. However, stance detection alone provides only a surface-level view of the argumentative landscape. Knowing that an article opposes gun control tells us the position, but not the reasoning behind it—whether the author argues from constitutional rights, public safety data, or personal liberty concerns.
\par
\cljournalquestion{What is the relationship between understanding the distribution of stances across a large collection of articles and capturing the full spectrum of opinions on controversial issues? What does "stance distribution" mean in this paper? What is the relationship between stance distribution and argument reasoning in the work? (Reviewers 2 \& 3)}

\begin{figure*}[h]
    \centering
    \includegraphics[width=1\textwidth]{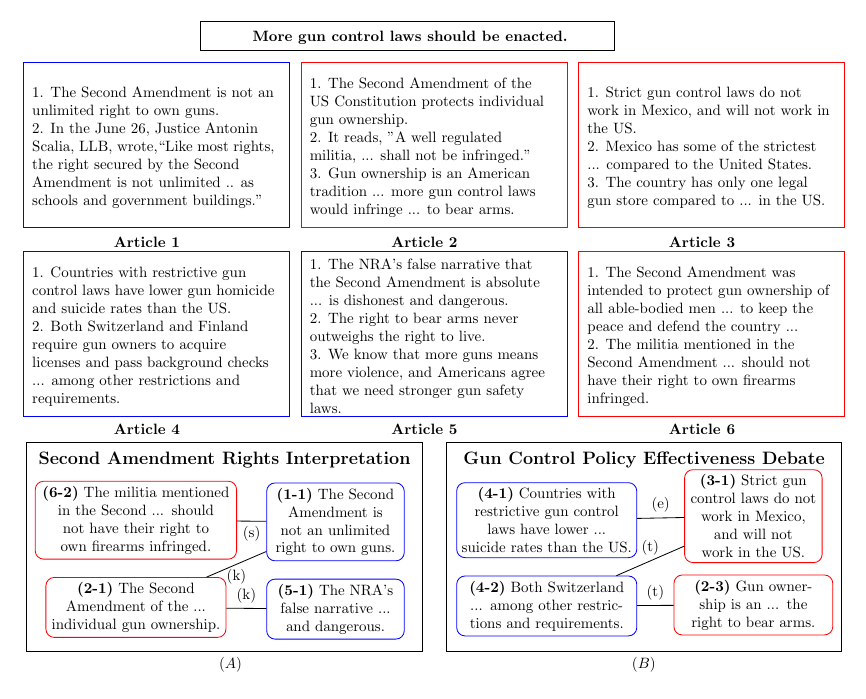}
    \caption{\textbf{top row -- } a sample claim, \textbf{next 2 rows -- } six articles associated with the claim, each article's stance is shown via the color of the box (blue refers to \textit{left} stance, red refers to \textit{right} stance), \textbf{bottom row -- } an illustrative output from our system that consolidates the arguments from the previous two rows into graph-based communities, each community is a Bipolar Bipartite Graph which reveals the controversial nature of argument interactions; The edge labels represent different types of relations among arguments: (e), (k), (s) and (t) are for Entity, Keyword, Semantic and Topic similarity edges respectively.
    \label{fig-sj-cma}
}
\end{figure*}

\cljournalanswer{Understanding the distribution of stances across a large collection of articles is crucial for capturing the full spectrum of opinions on controversial issues. However, traditional approaches that simply aggregate stance classifications provide only a coarse-grained view. For instance, knowing that "40\% of articles support and 60\% oppose gun control" fails to reveal the diverse reasoning frameworks underlying these positions. In this work, \textbf{stance distribution} refers to analyzing not merely the aggregate proportions of stances, but the distribution of argumentative perspectives—the distinct thematic reasoning frameworks through which authors approach a controversial claim. Stance distribution and argument reasoning are intrinsically connected. A comprehensive stance distribution analysis would reveal a nuanced landscape: some perspectives may exhibit consensus (homogeneous stance distribution, where arguments within a perspective largely agree), while others demonstrate polarization (heterogeneous distribution with competing viewpoints within the same reasoning framework). In Fig. \ref{fig-sj-cma}, we show six articles associated with the claim ``More gun control laws should be enacted", where articles 1, 4, and 5 take a \textit{left} stance (blue boxes) while 2, 3, and 6 take a \textit{right} stance (red boxes) w.r.t. the claim. A fine-grained analysis of this example reveals the perspectives:  articles 1, 2, 5, and 6 present arguments that focus on the implications of the Second Amendment in gun control laws; articles 2, 3, and 4 focus on the impact of gun control laws in different countries. This granular, perspective-level analysis is essential for exposing the full spectrum of opinions by surfacing both dominant and underrepresented narratives, potential biases in media coverage, and the specific aspects of an issue where agreement or disagreement concentrates. Only through such perspective-level stance distribution analysis can researchers gain deeper insights into the argumentative landscape, disrupt filter bubbles, and foster more informed decision-making on complex socio-political issues.}
\par
Recent research in argument mining and summarization shows that arguments across multiple articles often cluster around common themes or perspectives, even when written independently \citep{hasan2014you, trabelsi2019contrastive, trabelsi2019phaitv, quraishi2018viewpoint, jurkschat2022few}. Inspired by this finding, we study the interesting research question: ``\textbf{How can we design a theme-based argument organization system that could help users understand the stance distribution and the landscape of opinions on controversial topics?}'' Such a system could reveal both homogeneous (where authors largely agree) and heterogeneous (where significant disagreement exists) patterns within each thematic community while ensuring the arguments within each community are coherent, relevant to the theme, and informative.
\par

\cljournalquestion{What constitutes "socio-political issues" and what problems or challenges do they include? (Reviewer 3)}

\cljournalanswer{Analyzing stance distribution is particularly crucial for socio-political issues—controversial topics at the intersection of political ideology, public policy, and societal values where multiple stakeholder groups hold competing perspectives. Examples include gun control, abortion, immigration, healthcare policy, and climate change. These issues pose unique challenges that make understanding stance distribution especially complex.}

\cljournalanswer{First, arguments on socio-political issues invoke fundamentally different perspectives—the underlying reasoning themes that shape how stances are supported or opposed. For instance, in support of the claim "Abortion should be banned nationwide," an article might draw on the perspectives of \textit{life begins at conception} and \textit{contraceptive mandates}. Two authors taking the same stance may reason from entirely different perspectives: both may oppose a nationwide abortion ban, but one reasons from bodily autonomy ("A nationwide abortion ban would unjustly deny a woman's autonomy over her body") while another argues from public health consequences. Recognizing these perspectives is not trivial because they are rarely explicitly stated in articles, necessitating sophisticated semantic analysis to uncover the underlying reasoning themes. This implicit nature makes traditional keyword-based or surface-level analysis insufficient.}

\cljournalanswer{Second, perspectives are inherently context-dependent and dynamic. Two different discussions about the same claim may invoke entirely different sets of perspectives based on current events, cultural context, or the specific focus of the discourse. This dynamic nature makes it impractical to rely on predefined perspective categories \cite{bar2020arguments, jurkschat2022few} or existing datasets for Reason Classification \cite{hasan2014you}, as such approaches would inevitably make the model domain-dependent and fail to capture emergent perspectives.}

\cljournalanswer{Third, the complexity increases because (1) a single article may include both supporting and opposing arguments for a perspective, such as presenting conflicting views on \textit{contraceptive mandates} before stating a final stance on the claim, and (2) multiple articles may offer mutually opposing arguments on a shared perspective—one asserting "A nationwide abortion ban would safeguard the embryo's inherent right to life" with another contending "A nationwide abortion ban would unjustly deny a woman's autonomy over her body," both relevant to the perspective \textit{life begins at conception}.}

\cljournalanswer{Fourth, even after successfully partitioning arguments into distinct perspectives, each perspective might accumulate an overwhelming volume of arguments. The sheer volume of independently authored content makes it difficult for individuals to comprehend the full landscape of perspectives and identify areas of consensus versus conflict. Existing approaches to argument partitioning typically employ list-based presentations of clustered arguments \citep{dumani2021fine, trabelsi2019phaitv, alshomary2021key}. While this provides some organization, it fails to capture and communicate the \textit{relationships} between arguments, leaving users to struggle to understand why certain arguments are grouped or how they relate to each other. Additionally, these approaches often do not explicitly address the challenge of helping users understand opposing arguments. Varying writing styles, argument structures, and levels of expertise among authors further compound this challenge.}
\par
\cljournalquestion{Why do users need to understand why certain arguments are grouped or how they relate to each other in this task? (Reviewer 3)}

\cljournalanswer{Understanding argument groupings and their relationships is essential for several reasons. First, without seeing \textit{how} arguments connect—through shared entities, keywords, topics, or semantic meaning—users struggle to comprehend \textit{why} these arguments belong to the same perspective, reducing trust in the organization and limiting their ability to navigate complex debates. Second, relationships reveal the structure of reasoning. Third, in controversial debates, users need to see not just isolated opinions but the web of supporting and opposing arguments around shared themes to make informed judgments. For policymakers analyzing stakeholder positions, journalists identifying emerging narratives, or citizens seeking to understand multiple viewpoints, the \textit{connections} between arguments—not just their grouping—provide the critical context for synthesis and decision-making. Traditional list-based presentations obscure these relationships, leaving users with disconnected collections of statements rather than a coherent map of the argumentative landscape.}

\par

\par
\cljournalquestion{Why is the graph structure used in this work? (Reviewer 3)}

\cljournalanswer{To tackle these challenges, we introduce a graph-based approach to organize arguments into communities, making it easier to understand the complex interplay of perspectives. Our method starts by extracting key arguments from articles using rhetorical parsing \citep{saha2024stance}. Next, we build an interaction graph where nodes represent individual arguments, and edges capture various relationships between them, such as semantic similarity, shared topics, common keywords, or matching entities (e.g., "Second Amendment" or "US"). Using community detection and targeted graph operations, we group related arguments into meaningful communities, as illustrated in Fig. \ref{fig-sd-methodology}. Each community is an undirected graph where nodes are arguments, and edges show relationships like shared keywords or entities. These communities reveal different viewpoint distributions: single stance (all arguments align left, right, or center), dual stance (arguments split between left-right, left-center, or right-center), or mixed stance (arguments span left, right, and center). For example, Fig. \ref{fig-sj-cma} (bottom row) shows two argument communities. Community (A) focuses on interpretations of the Second Amendment regarding gun control laws. For instance, the arguments "The Second Amendment is not ... own guns" and "The Second Amendment of the US ... gun ownership" are connected by the shared keyword "Second Amendment." Community (B) explores the effectiveness of gun control policies across countries, linking arguments like "Countries with restrictive ... than the US." and "Strict gun control ... in the US." due to the common entity "US." Within these communities, we highlight nuanced interactions by focusing on a bipolar bipartite graph structure, where nodes are divided into two partitions representing opposing stances (e.g., left vs. right), and edges connect arguments across partitions when they share a strong semantic, topical, or entity-based relationship. This structure emphasizes key argumentative exchanges by tracing the longest stance-crossing path, allowing users to see how opposing arguments interact within a theme rather than viewing them in isolation.}
\par
\cljournalanswer{We adopt a graph-based structure because arguments across multiple independently authored articles form a complex web of relationships that cannot be adequately captured by linear lists or single-hierarchy structures. Arguments may relate to each other in multiple simultaneous ways: two arguments might share the same entity (e.g., "Second Amendment") while also exhibiting semantic similarity, and three arguments might cluster around a common sub-topic while presenting opposing stances. Graphs naturally model these multi-faceted relationships through typed edges, allowing us to preserve the rich interconnections that explain why arguments belong together. This graph-based methodology offers clear benefits over traditional list-based approaches, which present arguments as disconnected items within clusters without explaining their relationships. Lists may group arguments by similarity but fail to show \textit{why} they belong together or \textit{how} they interact—is it a shared entity? A common topic? Semantic alignment? In contrast, our graphs make relationships explicit through typed edges, allowing users to trace connections and understand the structure of perspectives. Furthermore, unlike tree-based hierarchies that enforce a single parent-child relationship, graphs accommodate arguments that relate to multiple other arguments in different ways, reflecting the true complexity of multi-article debates.}
\par
Our approach ensures that the desired qualities are attained at every step: rhetorical parsing selects the most informative arguments, the interaction graph captures coherent relationships within each theme, and community detection groups relevant arguments together, making the structure clear and meaningful. By modeling complex interactions between arguments across many articles, it provides comprehensive coverage while preserving important contextual relationships. Unlike simpler methods, our framework captures subtle opinion gradients and perspective variations by integrating arguments of varying stances within each community. Compared to cluster-based approaches, which may place arguments in a semantic graph space without clear connections, our method explicitly models relationships (e.g., semantic, topic, keyword, entity similarity), making the links between arguments transparent and understandable.

Our key contributions include:
\squishlist
    \item A graph-based framework for argument organization,  capturing rich relationships between arguments through multiple similarity measures
    \item A community detection approach that reveals both homogeneous and heterogeneous viewpoint distributions within argument clusters
    \item Graph simplification techniques, including a bipolar bipartite graph representation, that make complex argument relationships interpretable while preserving key information and highlighting direct interactions between opposing stances.
    \item An evaluation demonstrating our system's effectiveness in helping users understand diverse viewpoints through informative arguments
\squishend

%% file: related_works.tex
\section{Related Works}

The most relevant research w.r.t. our work has primarily evolved along three main directions: topic modeling, similarity-based argument clustering, and graph-based argumentation framework.
\paragraph{Topic modeling} Topic modeling approaches focus on uncovering latent viewpoints in contentious discussions. \citet{trabelsi2014finding,trabelsi2018unsupervised} propose a pipeline for detecting and clustering argument facets by leveraging author interactions in online debates. 
\citealp{trabelsi2019contrastive} advance this 
using a Phrase Author Interaction Topic-Viewpoint model, while \citealp{vilares2017detecting} propose a Bayesian approach. 
\citealp{misra2017using} 
focus on identifying central propositions and their underlying facets.
Several works  explore automatic extraction of key perspectives from documents \citep{vilares2017detecting, trabelsi2019contrastive, ibeke2017extracting}, 
leveraging LDA-like generative models with Gibbs Sampling for parameter estimation. For perspective labeling, while \citealp{trabelsi2019contrastive} uses phrase scoring and \citealp{ibeke2017extracting} treats it as a sentence ranking problem, these methods often lack deeper semantic understanding of perspectives and their contextual relationships.

\paragraph{Argument Clustering} Recent advances 
focus on grouping similar arguments and identifying representative viewpoints. \citet{shirafuji2021argument} studies 
automated key point generation, 
while \citet{lee2021argument} enhances clustering by combining multiple similarity matrices, and  \citet{reimers2019classification} demonstrates the effectiveness of contextualized embeddings (ELMo and BERT) in argument classification.
\citet{dumani2021fine} integrates polarity classification, frame classification, and meaning-based clustering, while \citet{conceiccao2022supporting} optimizes clustering of embeddings with dimensionality reduction. 
\citet{quraishi2018viewpoint} 
unearths viewpoints via graph partitioning of social interactions.  
Key point analysis has emerged as a crucial research direction \citep{bar2020arguments, eden2023welcome, bar2020quantitative, bar2021every, alshomary2021key, cattan2023key}. Although both supervised \citep{bar2020arguments} and unsupervised \citep{bar2020quantitative} approaches have been explored, they are highly reliant on manually curated data sets such as ArgKP, limiting their scalability to new topics. 
\par

Our work introduces a unified community-based approach that integrates semantic similarity and explicit argument relationships, offering a more comprehensive analysis of argument landscapes compared to traditional topic modeling or clustering methods. Unlike prior works that treat perspectives as latent variables or rely solely on argument similarity, we model nuanced interactions through a graph structure, capturing topical, semantic, entity-based, and keyword-based connections. Additionally, while existing methods \citep{trabelsi2019phaitv, jang2018explaining} present stance-segregated summaries, our interactive graph allows users to explore diverse opinions within each perspective, preserving rich interconnections and enabling more interpretable argument organization.

\paragraph{Graph-based structures for argumentation}
\rudrabluetext{Graph-based structures for argumentation, such as Abstract Argumentation Frameworks (AAFs) and the Argument Interchange Format (AIF), have significantly shaped computational argumentation, yet their scope differs markedly from our multi-article paradigm. AAFs, formalized by \citet{Dung1995}, are abstract in that they model arguments as atomic nodes without specifying their textual or semantic content, focusing solely on attack relations between them \citep{Baroni2007}. Extensions like value-based frameworks \citep{BenchCapon2003}, which introduce support relations (where arguments reinforce each other), and dependency graphs \citep{Bistarelli2023}, which assign weights to edges to reflect argument strength, enhance expressiveness but still require practitioners to manually map real-world arguments and relations, limiting scalability for content-rich corpora. The AIF, proposed by \citet{Chesnevar2006}, offers a standardized format with I-nodes (information) and S-nodes (relations), applied in tools like Araucaria for single-text annotation \citep{Reed2004}, dialogue modeling \citep{Reed2008}, and web-based argument aggregation \citep{Rahwan2007}. Recent work, such as \citet{Clayton2024}, introduces Argument Summary Graphs (ASGs) to summarize dialogical debates, with nodes as summarized key points and edges as support or attack relations inferred from conversational flow. These approaches are constrained to single-source settings—individual documents, dialogue threads, or web discussions—where edge relations are often manually annotated (e.g., in Araucaria) or derived from the source’s structure, such as reply links in dialogues or web forums, limiting perspective diversity to a unified context. While web content may include varied viewpoints, it remains bound to specific discussion threads. In contrast, our work constructs an interaction graph from a large corpus of independently written, expert-authored articles, automatically inferring content-driven edges (e.g., semantic, topic, keyword, entity) to capture inter-article relationships across diverse contexts (e.g., legal, ethical, social).}

%% file: problem_formulation.tex
\section{Problem Formulation} \label{problem-formulation} 
We formally define the problem of community-based argument organization and clarify our design choices. Given a collection of articles $\mathcal{A} = \{A_1, A_2, ..., A_n\}$ with their associated stances (for the choice of stance labels, please, see below) w.r.t. a topic/claim $T$, let $\mathcal{ARG}$ be the set of arguments
across all articles. 
\eat{Each article can adopt one of the three stances towards $T$: \textit{left}, \textit{right}, \textit{center}}
For simplicity, we assume that all arguments in an article adopt the same stance as the article. The objective is to partition and organize these arguments into a set of communities $\mathcal{C} = \{C_1, C_2, ..., C_m\}$, where the number of communities $m$ and the maximum community size, $k$ can optionally be specified by the user. Each community $C_j$ is represented as an undirected graph with arguments as vertices, with edges connecting related arguments. Edge labels indicate the type of similarity relationships, such as: topic, semantic, keyword, and entity. 
Each community is associated with a viewpoint. A viewpoint is determined based on the stances of the arguments that are part of the community. For example, suppose a community only has political arguments of `left' stance. In that case, the viewpoint will be `left aligned', whereas if it contains arguments of `left' and `right' stances, then the viewpoint will be `left and right aligned'. The resulting communities should provide a comprehensive and interpretable landscape of viewpoints, enabling users to explore nuanced argument interactions and uncover both homogeneous and heterogeneous perspective distributions within each community.

\subsection{Design Choices}\label{design-choices}

\textbf{Influence of Stance Label Selection on Argument Organization}: Stance labels categorize the positions articles take relative to a topic or claim. For example, in political debates, labels might be left, right, or neutral, reflecting ideological positions. In other domains, such as policy or product discussions, labels like pro, con, or neutral are often used to indicate support, opposition, or impartiality. \LL{The stance labels we use for each dataset follow the conventions used by the curators of the datasets}\eat{reflect the conventions used in the datasets we tested}. This choice is an \textit{inherent property} of the datasets and made by the curators of those datasets. For instance, the Allsides dataset, designed around political discourse, employs left, right, and center, while the Perspectrum Dataset uses support and undermine, with its articles collected from idebate.com, debatewise.org, and procon.org. Other examples include the Createdebate dataset \cite{saha2024stance}, which uses pro and con for debate-focused arguments, and the Semeval-2016 Task 6 dataset \cite{mohammad-etal-2016-semeval}, which applies support, against, and neutral labels for stance detection in social media. For datasets with only pro and con labels, such as those from procon.org, we will not have a mixed community, as mixed communities typically include arguments from three stance classes (e.g., left, right, and neutral, or pro, con, and neutral). We note that the specific choice of stance labels is orthogonal to our core methodology.
\newline

\textbf{Capturing Intra- and Inter-Article Discourse Relations for Argument Organization}: Our work utilizes the “Stance Tree” architecture \cite{saha2024stance}, which captures the discourse dependency among arguments in an article. In this work, we create an Interaction graph, which preserves the edges that occur in an intra-article analysis. However, since our focus is on analyzing stance distribution on a large collection of articles, we aim to investigate the presence of semantic relations that we expect in an inter-article analysis (rather than low level intra-article discourse relations). \LL{We capture these relations through the edges in a community and the relations can be semantic, shared-entity, shared-topic, or shared-keyword.} As for argumentation relations such as `support' and `attack', we capture `attack' relations through the edges of a bipolar bipartite graph, where an edge connects arguments with opposing stances. Similarly, for unipolar communities, where the majority of arguments belong to a single stance, the edges represent a `support' relation (for more details, please see Section \ref{Budget-Constrained Optimization}).

%% file: methodology.tex
\section{Methodology}
\begin{figure*}[htbp]
    \centering
    \includegraphics[width=1\textwidth]{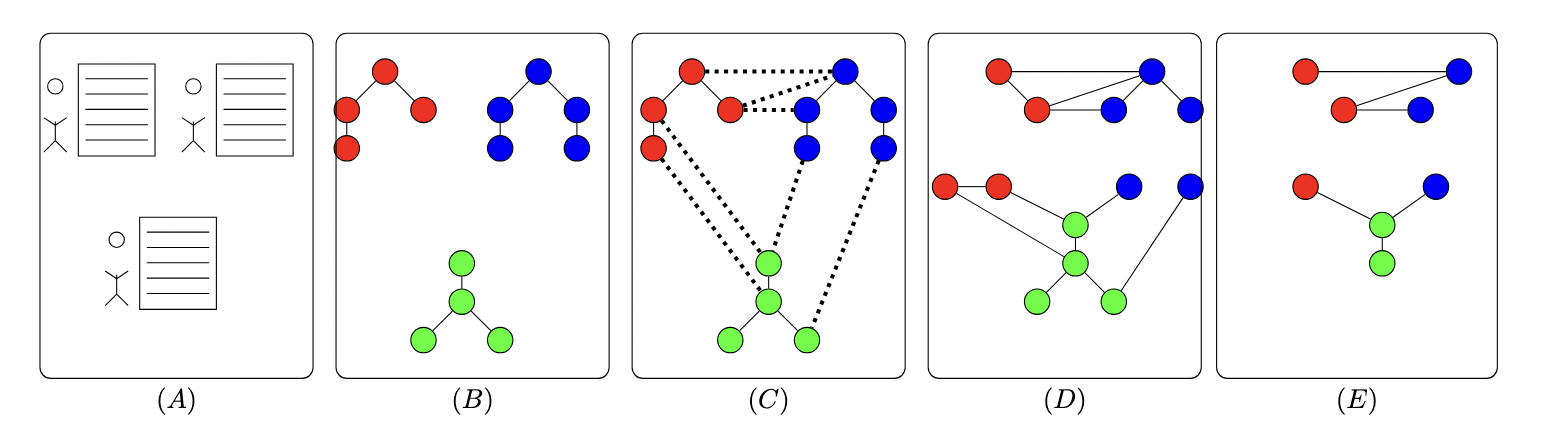}
    \caption{\textbf{(A)} A collection of articles w.r.t. a topic, (\textbf{B})-(\textbf{E}) A step-by-step overview of our methodology, \textbf{(B)} Stance Trees constructed for each article (blue, red, and green colors are assigned to left, right, and center stances respectively), \textbf{(C)} An Interaction graph built on top of the Stance Trees, with dotted edges reflecting different relations between an argument pair, \textbf{(D)} Communities detected from the Interaction Graph, \textbf{(E)} Graph Simplification Stage: the top graph reflects a Bipolar community, the bottom graph reflects a Mixed community.}
    \label{fig-sd-methodology}
\end{figure*}

Our approach, STIC, consists of three main phases: (1) argument extraction and \underline{S}tance \underline{T}ree creation, (2) \underline{I}nteraction graph construction and enhancement, and (3) \underline{C}ommunity detection and analysis. Figure~\ref{fig-sd-methodology} presents an overview of our methodology. For each claim-article pair, first, we create a discourse dependency tree following the work of \cite{saha2024stance} (Figure \ref{fig-sd-methodology} (B)). Since arguments from different articles often share a common theme, we then construct an Interaction graph by adding edges between arguments if they share some common properties (Figure \ref{fig-sd-methodology} (C)). Next, we detect coherent communities from the Interaction graph by applying a Community Detection algorithm (Figure \ref{fig-sd-methodology} (D)). Finally, we simplify the graph to create digestible outputs for end-users while retaining the crucial information from each community as well as showing interesting properties such as homogeneity and heterogeneity between opinions about controversial topics (Figure \ref{fig-sd-methodology} (E)).

\subsection{Argument Extraction and Stance Tree Creation}
Following \citet{saha2024stance}, we leverage Discourse Segmentation Theory to extract and organize arguments from input articles in an unsupervised manner, constructing a rhetorical parsing tree, termed a Stance Tree $ST_i = (V_i, E_i)$, for each article $A_i \in \mathcal{A}$. Here, $V_i$ represents the set of arguments extracted as sentence-level units from article $A_i$, and $E_i$ denotes rhetorical relationships (e.g., elaboration, justification) derived from Rhetorical Structure Theory (RST) \citep{mann1988rhetorical}, capturing the hierarchical argumentative structure. In that work, the authors transformed the RST-based Discourse Parse Tree (DPT) into a Discourse Dependency Tree (DDT) to map dependencies between Elementary Discourse Units (EDUs) or sentences, creating a compact representation of argumentative flow \citep{li2014text,hirao2013single} while a pruning module discards irrelevant arguments. Arguments closer to the root are prioritized as they reflect the article’s central claim, supported by subordinate arguments, ensuring the most salient points are identified without \LL{the need for} manual annotation.
\par
In prior work \citet{saha2024stance}, an external stance predictor model was used to predict the stance of each node in the DDT i.e. the arguments w.r.t. the claim and later, they were aggregated using the Dempster Shafer Theory \cite{shafer1992dempster}. In this work, we simplify the Stance Tree construction by assigning each argument the article-level stance label. This allows us to focus on inter-article argument organization and stance distribution rather than intra-article stance prediction. Also, we replace the edge labels that would denote the intra-DDT discourse relations with a generic type `stance' as explained in Section \ref{design-choices}. This structure generates explainable outputs by highlighting pivotal arguments near the root, which we use in our community-based approach to construct an Interaction graph across multiple articles, linking arguments via various similarity based relationships to form perspective-driven communities.

\subsection{Interaction Graph Construction}
We construct a global interaction graph $G = (V, E)$ by merging Stance Trees from all articles and augmenting them with additional edges capturing topic, semantic, keyword, and entity-based relationships. Formally, let $V = \bigcup_{i=1}^n V_i$ be the union of all argument nodes across articles, and $E = E_{base} \cup E_{topic} \cup E_{sem} \cup E_{key} \cup E_{ent}$, where $E_{base} = \bigcup_{i=1}^n E_i$ represents the original Stance Tree edges. These additional edge types enhance the graph by connecting arguments across articles, enabling the discovery of inter-article relationships critical for community formation. \footnote{It's possible for two arguments to be similar in terms of multiple relationships, for example: both topic and semantic. In such cases, we give priority to edges in the order of topic, similarity, keyword, and entity.}

\paragraph{\textbf{Topic Similarity Edges}}
Arguments under the same topic often share common sub-topics, revealing nuanced perspectives. Consider these arguments on ``gun control'':
\begin{itemize}
    \item \textbf{Argument 1}: The Second Amendment guarantees Americans' fundamental right to bear arms as a crucial safeguard against government overreach and a means of self-defense.
    \item \textbf{Argument 2}: The Second Amendment doesn't guarantee unrestricted individual gun ownership in modern society.
\end{itemize}
Both arguments discuss the Second Amendment (sub-topic), representing contrasting interpretations. We utilize BERTopic \citep{grootendorst2022bertopic} to identify sub-topics by clustering argument embeddings, adding an edge $(arg_i, arg_j) \in E_{topic}$ if $topic(arg_i) = topic(arg_j)$. BERTopic employs HDBSCAN for clustering and c-TF-IDF for topic representation, with parameters tuned to balance granularity and coherence (details in Section \ref{experimental-setup}). These edges help create coherent communities while also highlighting diverse arguments within subtopics, aiding users in understanding nuanced perspectives.

\paragraph{\textbf{Semantic Similarity Edges}}
Topic similarity alone may miss arguments with nuanced but related reasoning. Consider these education-related arguments:
\begin{itemize}
    \item \textbf{Argument 1, topic: Technology in Education}: \rudrabluetext{Giving students access to modern technology in classrooms empowers them to engage with learning in more interactive and meaningful ways.}
    \item \textbf{Argument 2, topic: Curriculum Choice}: \rudrabluetext{Allowing students more freedom in choosing their subjects helps them take ownership of their education and stay motivated throughout their studies.}
\end{itemize}
\laks{how is argument 2 relevant to immigration?}\rudraorange{Can you check the arguments above? I have replaced the previous arguments with "Education" topic-related arguments.}
\eat{Both of these arguments critique a policy (permits, compliance rules) for creating delays and costs, ultimately harming a vulnerable group (low-income families, rural communities) by limiting access to a critical resource (housing, clean power).}

Both sentences follow a similar structure ("Allowing/Giving students X helps/empowers them to Y"), use similar vocabulary, and focus on improving student engagement. We compute semantic similarity using a Sentence-Transformer model \citep{reimers-2019-sentence-bert}, adding an edge $(arg_i, arg_j) \in E_{sem}$ if their cosine similarity $\text{sim}_s(arg_i, arg_j) > \tau_{sem}$, where $\tau_{sem} = 0.7$ (tuned empirically). This approach captures implicit logical relationships, bridges diverse reasoning types (e.g., emotional, economic), and enriches community structures by connecting arguments with semantically similar conclusions.

\paragraph{\textbf{Keyword Edges}}
Shared keywords often indicate related arguments, revealing thematic connections. Consider these arguments on abortion:
\begin{itemize}
    \item \textbf{Argument 1}: Promoting \textbf{contraceptive} education and access in schools helps young people make informed decisions about their reproductive health.
    \item \textbf{Argument 2}: The failure rate of \textbf{contraceptive} methods ... demonstrates that unwanted pregnancies can still occur and women need access to abortion as a backup option.
\end{itemize}
The shared keyword ``contraceptive'' links preventive and fallback perspectives. Using KeyBERT \citep{grootendorst2020keybert}, we extract key phrases $K(arg_i)$ with confidence scores $s(k) \geq \theta = 0.5$ (empirically set) and add an edge $(arg_i, arg_j) \in E_{key}$ if $K(arg_i) \cap K(arg_j) \neq \emptyset$. KeyBERT leverages BERT embeddings and Maximal Marginal Relevance for diversity, ensuring robust phrase extraction (details in Section \ref{experimental-setup}). These edges highlight how terms support different positions, trace concept evolution, and enrich communities \LL{to help navigate }  discussions about controversial topics.

\paragraph{\textbf{Entity Edges}}\label{entity-edges}
Arguments sharing common entities (e.g., politicians, organizations, locations) often provide complementary perspectives. Consider these foreign policy arguments:\\
\textbf{Argument 1}: In public and reportedly in private, \textbf{Harris} has demonstrated far more concern with the suffering of Palestinian civilians.\\
\textbf{Argument 2}: While some coverage ... a foreign-policy neophyte, \textbf{Kamala Harris} would come into office with more global experience than Bill ...\par
In the example above, two arguments discussing Kamala Harris—one highlighting her concern for Palestinian civilians and another emphasizing her foreign policy experience—offer different insights into her role. We connect arguments if they share at least one common entity. Entity-based edges track how entities are discussed across contexts, reveal temporal or contextual relationships, and help build coherent narratives around key figures or institutions.


\begin{figure}[p]
  \centering
  
  \begin{subfigure}{\textwidth}
    \centering
    \includegraphics[width=0.7\textwidth]{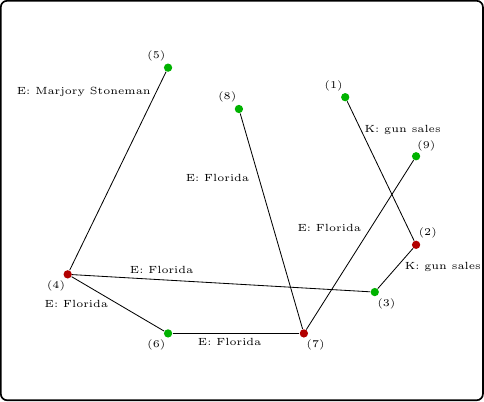}
    \caption{A sample community from \ourapproachroberta{} on the topic of ``Gun Control''; it contains edges of type \textbf{Keyword} and \textbf{Entity}, labeled as K: (e.g., K: gun sales) and E: (e.g., E: Florida) in the figure. Green and Red indicate arguments with center and right stances, respectively. There is a \textbf{Keyword} edge labeled "gun sales" connecting argument \textcolor{green}{1} to argument \textcolor{red}{2}, and another \textbf{Keyword} edge with the same label between argument \textcolor{red}{2} and argument \textcolor{green}{3}. Additionally, there are \textbf{Entity} edges labeled "Florida" linking argument \textcolor{green}{3} to argument \textcolor{red}{4}, as well as argument \textcolor{red}{4} to argument \textcolor{green}{5}. \laks{you shd probably nu,ber the arguments in the visualization directly. directly dereferencing argument numb ers against dots in the figure is confusing.}\rudraorange{please, see the updated caption.}
}
    \label{fig:stic-r-survey-output-first-part}
  \end{subfigure}
  
  \vspace{0.5cm} 
  
  \begin{subfigure}{\textwidth}
    \centering
    \resizebox{\textwidth}{!}{%
    \begin{tikzpicture}
      \draw[thick] (0,0) rectangle (16, -10);
      \node[anchor=north west, align=left, text width=15cm] at (0.5, -0.5) {
        \textbf{Community 2} \\
        \vspace{0.3cm}
        \textbf{Viewpoint:} Right and Center-aligned \\
        \vspace{0.3cm}
        \textbf{Perspective Label:} National Debate on gun control and gun rights \\
        \vspace{0.3cm}
        \setcounter{enumi}{0} 
        \begin{enumerate}[label={\color{blue}\arabic*.}]
          \greenitem (Article Id: 49) Florida lawmakers, spurred by last month's deadly high school shooting, gave final passage on Wednesday to a bill to raise the legal age for buying rifles, impose a three-day waiting period on all gun sales and allow the arming of some school employees.
          \reditem (Article Id: 28) It would also raise significant constitutional questions that gun Rights groups say they will challenge in court if lawmakers attempt to curb gun sales.
          \greenitem (Article Id: 15) Two major US retailers have announced new restrictions on gun sales following the shooting at a Florida school where 17 people died.
          \reditem (Article Id: 39) They face growing pressure after the deadly Feb. 14 shooting at Parkland, Florida's Marjory Stoneman Douglas High School that killed 17 people.
          \greenitem (Article Id: 5) NRA chief executive Wayne LaPierre echoed President Donald Trump's call to arm teachers to prevent school shootings, and weighed in on a long-running political and cultural divide over access to weapons that has been inflamed by last week's massacre at a Florida Marjory Stoneman high school that killed 17 students and staff.
        \end{enumerate}
      };
    \end{tikzpicture}%
    } 
    \caption{An abbreviated list of arguments from the top figure.}
  \end{subfigure}
  
  \caption{An abbreviated output from \ourapproachroberta{} on the topic of ``Gun Control'', as shown to turkers during the survey (more details in Section \ref{ref:qualitative-Analysis})}
  \label{fig:stic-r-survey-output}
\end{figure}

\subsection{Community Detection and Analysis}\label{Budget-Constrained Optimization}
We apply the Eva algorithm \citep{citraro2020eva,citraro2020identifying} to the enriched interaction graph $G$ to detect communities $\mathcal{C} = \{C_1, C_2, \ldots, C_m\}$. The Eva algorithm extends the Louvain method by incorporating node attributes, optimizing a linear combination of Newman’s modularity (for structural cohesion) and purity (based on the frequency of dominant stance labels within communities). A parameter $\alpha \in [0,1]$ (set to 0.5 for balanced optimization) tunes the trade-off between structural and attribute-based clustering, ensuring communities reflect both argumentative connections and stance coherence (details in Section \ref{experimental-setup}). For each community $C_j$, we construct a \rudrabluetext{weighted} Minimum Spanning Tree (MST) $T_j = (V_j, E^{MST}_j)$ using Kruskal’s algorithm \citep{kruskal1956shortest} to capture essential relationships while minimizing edge complexity. 
\par
The MST structure reveals three community patterns:
\begin{itemize}
    \item \textbf{Homogeneous Communities}: Dominated by a single stance, forming uniform structures where edges primarily connect arguments of the same stance, highlighting cohesive viewpoints.
    \item \textbf{Bipolar Communities}: Featuring two dominant stances, forming bipartite graphs that highlight contentious argument paths. Unlike existing methods relying on topic-level controversy \citep{benslimane2023explaining}, reader discussions \citep{beelen2017detecting}, or threaded social media debates \citep{li2023novel,guimaraes2021x,zhong2020integrating,koncar2021analysis}, our approach detects controversy directly from opposition patterns in the graph without requiring training data or threaded discussions, leveraging natural stance conflicts across independent articles.
    \item \textbf{Mixed Communities}: Containing diverse stances, resulting in heterogeneous structures that connect arguments from multiple perspectives, facilitating exploration of varied viewpoints.
\end{itemize}

Edge weights in the MST are assigned based on stance alignment: $w(arg_i, arg_j) = 1$ for same-stance edges and $w(arg_i, arg_j) = 0.5$ for opposing-stance edges, prioritizing diverse stance representation. \laks{what does this weight assignment mean given that we seem to be interested in identifying communities with different "interesting" stance distributions, e.g., homogeneous, polarized/contrasting, mixed? what sort of impact would a different weight assignment have on the quality of communities extracted?} 
\rudrabluetext{This weight assignment reflects our emphasis on identifying communities exhibiting controversial or heterogeneous structures. By assigning lower weights to opposing stance edges ($w = 0.5$) and higher weights to same stance edges ($w = 1$), the construction of the MST prioritizes the inclusion of cross-stance edges, thus favoring paths that reveal contrasting viewpoints. This encourages the emergence of bipolar or mixed communities, supporting our objective of surfacing latent stance-driven controversy without relying on external supervision or conversation structure. Alternative weighting schemes would shift the nature of the extracted communities. For instance, assigning lower weights to same-stance edges would lead the MST to favor homogeneous structures, potentially reinforcing ideologically cohesive clusters while overlooking internal disagreement. A uniform weighting scheme, in contrast, would reduce sensitivity to stance alignment, resulting in MSTs that reflect connectivity alone. Thus, the proposed stance-sensitive weighting strategy plays a critical role in guiding the community structure toward configurations that reflect meaningful stance interplay and argumentative contrast.}

To handle user-specified constraints on community size ($k$) and count ($m$), we implement two strategies:
\paragraph{Argument Selection ($k$ constraint)} We propose two methods to reduce communities to $k$ arguments:
\begin{itemize}
    \item \textbf{Degree-based Filtering}: Iteratively removes nodes with the lowest degree (i.e., least connected) until the community size reaches $k$, preserving the most interconnected arguments.
    \item \textbf{Quality-based Filtering}: Uses a RoBERTa-based \citep{liu2019roberta} argument quality scorer fine-tuned on IBM-Rank-30k \citep{gretz2020large} to assign quality scores to arguments, iteratively removing the lowest-scoring arguments until size $k$ is reached. If multiple components arise, we retain the largest connected component to maintain coherence.
\end{itemize}

\paragraph{Community Selection ($m$ constraint)} 
 \laks{i checked the whole paper: it seems to be the first time \ourapproachroberta{} is mentioned. the reader will have no idea what it is.}
 \rudrabluetext{The Degree-based and Quality-based filtering approaches are referred to as \ourapproachmst{} and \ourapproachroberta{}, respectively. Figure~\ref{fig:stic-r-survey-output} presents an example output of a community from \ourapproachroberta{} for the topic of ``Gun Control.''} We compute community quality as the mean argument quality score using the fine-tuned RoBERTa model, selecting the top $m$ communities to prioritize high-quality perspectives. For each community, we generate a descriptive perspective label using OpenAI’s \texttt{GPT-3.5-turbo-instruct} model in a few-shot setting. The model is prompted with examples of argument collections paired with perspective labels (e.g., “Arguments supporting stricter gun control” for a homogeneous pro-control community), ensuring labels accurately reflect the collective meaning, capture stance diversity, and provide intuitive entry points for users.
An example of the output from our approach is shown in Figure \ref{fig:stic-r-survey-output}, where the perspective label is `National debate on gun control and gun rights'. Overall, this multi-stage pipeline preserves meaningful argument relationships and facilitates navigation of diverse viewpoints, addressing limitations of list-based approaches by explaining why arguments belong to a community and simplifying large argument sets \LL{facilitating} user comprehension.

%% file: experiment.tex
\section{Experiments}

\rudraorange{Below, I added a paragraph to explain the goal of the experiments as well as the outline.}
\rudrabluetext{We evaluate the effectiveness of our community-based argument organization system on large-scale datasets comprising independently authored articles. By incorporating a large number of articles, we aim to test the system's \laks{not sure robustness is the right term. i edited it slightly. see if you like it.} ability to handle complex, real-world scenarios that involve a wide range of perspectives. We also seek to ensure broad coverage of topics and to capture subtle differences in stance. The goal is to assess the system’s ability to generate coherent, informative, and relevant argument communities that reflect these diverse perspectives. Our evaluation includes both quantitative analysis, using automated metrics for coherence, informativeness, and relevance, and qualitative analysis through a human study designed to capture nuanced user judgments. This section is organized as follows: Sections \ref{dataset} and \ref{baselines} describe the datasets and baselines; Section \ref{experimental-setup} outlines the experimental setup; Sections \ref{quantitative-analysis}, \ref{baseline-comparisons}, \ref{metric-wise-performance-analysis}, and \ref{detailed-analysis-allsides-dataset} report the quantitative results; Sections \ref{ref:qualitative-Analysis}, \ref{human-study-results}, and \ref{comparison-between-baselines} present the qualitative findings; and Section \ref{comparison-between-quantitative-and-qualitative-results} offers an comparison between the quantitative and qualitative assessment of system performance.}

\begin{table*}[t]
\centering
\small
\begin{tabular}{cccccp{3cm}}
\toprule
\textbf{Dataset} & \textbf{Topic} & \textbf{\#Articles} & \textbf{Avg. Sent.} & \textbf{Avg. Words} & \textbf{Stance Distribution(\%)} \\
\midrule
Allsides & Education & 200 & 49.91 & 1,188.57 & L(30.5) C(30.0) R(39.5) \\
& Elections & 200 & 41.33 & 1,040.67 & L(46.0) C(25.5) R(28.5) \\
& Foreign Policy & 200 & 37.34 & 990.10 & L(30.5) C(32.5) R(37.0) \\
& Gun Control & 200 & 36.26 & 927.62 & L(27.0) C(22.5) R(50.5) \\
& Healthcare & 200 & 37.87 & 950.03 & L(36.0) C(20.5) R(43.5) \\
& Immigration & 200 & 38.84 & 1,031.85 & L(32.5) C(22.0) R(45.5) \\
& National Security & 200 & 43.54 & 1,177.60 & L(23.0) C(39.0) R(38.0) \\
& Politics & 200 & 45.84 & 1,135.25 & L(32.5) C(29.0) R(38.5) \\
& Supreme Court & 200 & 38.88 & 1,053.60 & L(31.5) C(27.0) R(41.5) \\
& Terrorism & 200 & 44.81 & 1,115.34 & L(40.5) C(25.5) R(34.0) \\
\midrule
Perspectrum & \parbox[t]{2.5cm}{Abolish nuclear weapons.} & 54 & 8.04 & 208.43 & S(50.0) U(50.0) \\[2pt]
& \parbox[t]{2.5cm}{All nations have a right to nuclear weapons.} & 55 & 7.95 & 206.47 & S(49.1) U(50.9) \\[2pt]
& \parbox[t]{2.5cm}{Social networking sites are good for our society.} & 60 & 5.25 & 131.20 & S(55.0) U(45.0) \\[2pt]
& \parbox[t]{2.5cm}{There is a need for developing tactical nuclear weapons.} & 54 & 8.04 & 208.81 & S(48.1) U(51.9) \\
\bottomrule
\end{tabular}
\caption{Analysis of articles from Allsides and Perspectrum datasets. For Allsides, L/C/R represents Left/Center/Right stance percentages. For Perspectrum, S/U represents Support/Undermine stance percentages. Average sentence and word counts are averaged over articles per topic}
\label{tab:dataset-analysis}
\end{table*}

\subsection{Datasets}\label{dataset}

\LL{We conduct our experiments on two major publicly available datasets} -- \href{https://github.com/ramybaly/Article-Bias-Prediction}{Allsides} and \href{https://github.com/CogComp/perspectrum}{Perspectrum}. Table ~\ref{tab:dataset-analysis} reports their article counts, average sentence and word counts per topic (computed using Python's NLTK library), and stance distributions at the article level.
\newline 
\textbf{Allsides}: \rudrabluetext{The AllSides balanced news dataset \cite{baly2020we} contains expert-selected U.S. news articles from sources with different political orientations (left, right, center), which often exhibit spin bias, slant, and other forms of non-neutral reporting on political news. We chose a test set of 10 topics (Education, Elections, Foreign Policy, Gun Control, Healthcare, Immigration, National Security, Politics, Supreme Court, Terrorism) to ensure a comprehensive evaluation, each containing at least 200 articles. Each article is annotated with stance labels: left, right, or center.} 
\laks{1.list the datasets in alphabetical order unless there is some other logic (which i can't tel) to the way you list them . 2. the topics i can see in table 1 are not quite the same as what are listed here. 3. what was the criteria for selecting the topics for eval vis a vis those fo hyper parameter tuning?}\newline
\rudraorange{1. Resolved, 2. They are the same. 3. I have moved which topics were used for hyper-tuning to Section \ref{experimental-setup}.}
\newline
\textbf{Perspectrum}: Perspectrum \cite{chen2018perspectives} is a dataset of claims, perspectives, and evidence, making use of online debate websites to create the initial data collection, and augmenting it using search engines to expand and diversify the content. We picked four topics from the Perspectrum dataset that each have 50 or more articles\footnote{Perspectrum uses the stance labels \textit{support, undermine}}.  The topics are: `Abolish nuclear weapons.', `All nations have a right to nuclear weapons,' `Social networking sites are good for our society.', `There is a need for developing tactical nuclear weapons.'. 
\laks{were any of these used for HP tuning as well?}
\rudraorange{Short answer: no. I have moved which topics were used for hyper-tuning to Section \ref{experimental-setup}.}

\subsection{Baselines} \label{baselines}

\laks{how are the baselines related to \ourapproachroberta{}?} \rudraorange{Please, see the paragraph below in blue.}

\rudrabluetext{Our approach generates a concise, graph-based overview of arguments from a large collection of articles, organized by distinct perspectives within an interaction graph (see Figures 2 and 3). Unlike existing methods, it produces a set of graphs that visually and structurally represent argument stances and their relationships, rather than a textual summary. To our knowledge, no graph-based approaches for argument organization with a published codebase exist. The closest baselines are list-based and key-point-based methods, which organize arguments into clusters or match them to key points, typically represented as lists labeled by frequent words or phrases. These methods align closely with our goal of organizing arguments across multiple articles by grouping related perspectives, despite their non-graph-based output.} We compare our approach with three  baselines\footnote{Several other potential baselines were considered \cite{bar2020quantitative, benslimane2023explaining, gupta2022right, wei2021graph, lee2021argument, jang2018explaining, trabelsi2019phaitv}, but excluded due to code inaccessibility or communication challenges.}:

\begin{itemize}
    \item \textbf{\baselinei} (\citet{reimers2019classification}): A contextual word embedding approach designed to enhance argument clustering performance on the AFS dataset \citep{misra2017using}. This method produces clusters of arguments, with each cluster represented as a list.
    
    \item \textbf{\baselineii} (\citet{alshomary2021key}): A graph-based extractive summarization framework that employs a Siamese neural network architecture for argument-to-keypoint matching. It generates clusters of arguments, each presented as a list, with the key point serving as a perspective label. Figure \ref{fig:baselineii-survey-output} presents an example output of a community from \baselineii{} for the topic of ``Gun Control.''
    
    \item \textbf{\baselineiii} (\citet{Meer2024AnEA}): An LLM-based approach leveraging ChatGPT for key point generation and matching. The output consists of clusters of arguments, each displayed as a list, with the key point acting as a perspective label.  Figure \ref{fig:baselineiii-survey-output} presents an example output of a community from \baselineiii{} for the topic of ``Gun Control.''
\end{itemize}
\par

\begin{figure}[p]
  \centering
  \begin{subfigure}{\textwidth}
    \centering
    \begin{tikzpicture}
      \draw[thick] (0,0) rectangle (13.5, -8.5);
      \node[anchor=north west, align=left, text width=12.8cm] at (0.5, -0.5) {
        \textbf{Community 2} \\
        \vspace{0.2cm}
        \textbf{Viewpoint:} Left-aligned \\
        \vspace{0.2cm}
        \textbf{Perspective Label:} After a mass shooting four years ago, the state passed a new law where courts could be asked to temporarily bar an at-risk person from owning guns. \\
        \vspace{0.2cm}
        \textcolor{blue}{1.} (Article Id: 421) Slightly more than half of Texas registered voters say gun control laws should be stricter; a strong majority would require background checks for all gun purchases ...\newline
        \textcolor{blue}{2.} (Article Id: 372) After a mass shooting four years ago, the state passed a new law where courts could be asked to temporarily bar an at-risk person from owning guns. \newline
        \textcolor{blue}{3.} (Article Id: 372) A California law designed to help police ... might have stopped the shooter who killed 12 people at a country and western bar. \newline
        \textcolor{blue}{4.} (Article Id: 381) US court upholds ban on buying guns for medical-marijuana users. Gun purchases are off limits in the U. S. to anyone who uses medical marijuana or holds a state-approved medicinal marijuana card ... \newline
        \textcolor{blue}{5.} (Article Id: 346) Blumenthal said it is illegal to sell either guns or ammunition to certain groups of people including felons, the mentally ill and those who commit domestic violence.
      };
    \end{tikzpicture}
    \caption{An abbreviated output from \textbf{\baselineii{}} on the topic of ``Gun Control''; Blue indicates arguments with left stance.}
    \label{fig:baselineii-survey-output}
  \end{subfigure}

  \vspace{0.5cm} 

  \begin{subfigure}{\textwidth}
    \centering
    \begin{tikzpicture}
      \draw[thick] (0,0) rectangle (13.5, -10);
      \node[anchor=north west, align=left, text width=12.8cm] at (0.5, -0.5) {
        \textbf{Community 3} \\
        \vspace{0.2cm}
        \textbf{Viewpoint:} Center-aligned \\
        \vspace{0.2cm}
        \textbf{Perspective Label:} Raising the legal age for buying rifles and imposing waiting periods can help prevent impulsive acts of violence and reduce the likelihood of mass shootings. \\
        \vspace{0.2cm}
        \textcolor{green}{1.} (Article Id: 436) Americans have a constitutional right to openly carry firearms, a federal appeals court ruled Tuesday, delivering a major victory to gun rights supporters.\newline
        \textcolor{green}{2.} (Article Id: 440) One day after President Barack Obama won re-election, his Administration agreed to a new round of international negotiations to revive a United Nations-sponsored treaty regulating the international sale of conventional arms, which critics fear could affect the Constitutionally protected right of U. S. citizens to purchase and bear firearms. \newline
        \textcolor{green}{3.} (Article Id: 328) NRA chief executive Wayne LaPierre echoed President Donald Trump's call to arm teachers to prevent school shootings, and weighed in on a long-running political and cultural divide ...\newline
        \textcolor{green}{4.} (Article Id: 371) - Florida lawmakers, spurred by last month's deadly high school shooting, gave final passage on Wednesday to a bill to raise the legal age for buying rifles ....\newline
        \textcolor{green}{5.} (Article Id: 406) But if you are mentally ill, whether you're a veteran or not, just like if you're a felon, if you're a veteran or not, and you have been judged to be mentally infirm, you should not have a gun.
      };
    \end{tikzpicture}
    \caption{An abbreviated output from \textbf{\baselineiii{}} on the topic of ``Gun Control''; Green indicates arguments with center stance.}
    \label{fig:baselineiii-survey-output}
  \end{subfigure}
  \caption{Abbreviated outputs from \textbf{\baselineii{}} and \textbf{\baselineiii{}} on the topic of ``Gun Control'', as shown to turkers during the survey (more details in Section~\ref{ref:qualitative-Analysis}).}
  \label{fig:combined-baseline-survey-output}
\end{figure}

\subsection{Experimental Setup}\label{experimental-setup}

\laks{the description in this section is lacking context. for instance, even the datasets have not been introduced and yet the implementation details refer to different data sets. also "implementation details" doesn't sound like the right title for this section.}
\rudraorange{I have changed the subsection name from `Implementation Details' to `Experimental Setup'. I have moved the dataset descriptions to the beginning of the Experiment section.}
\rudrabluetext{For each dataset, we remove duplicate arguments from the article collection using Python's NLTK library, retaining only one instance when the Jaccard similarity between arguments is greater than or equal to 0.9. For hyperparameter tuning, we selected 5 topics with 200 articles from the Allsides dataset (Abortion, Gun Control, Election, Immigration, Taxes). For topics overlapping between the development and the test set, we selected 200 entirely new articles to prevent data leakage.} For the Allsides dataset, with Stance trees restricted to a depth of 2, yielding more than 700 arguments. For experiments on the Perspectrum dataset, given the shorter length of the articles compared to the Allsides, we do not pose a depth restriction on the Stance Tree. 
\par
For topic modeling, we perform offline training of the BERTopic model using 150 articles per topic collected via Python's Google search library. To enhance topic detection granularity, we segment these articles into 5-sentence paragraphs and filter out those not containing the topic. The BERTopic implementation uses UMAP initialization with 15 neighbors and five components, with the embedding model set to `paraphrase-MiniLM-L3-v2' and a topic size of 10. The model training occurs once per topic, and during inference, we average the results of 5 runs, assigning sub-topics only when confidence exceeds 70\%. \par
For semantic similarity computation, we employ SentenceTransformer's \textit{stsb-mpnet-base-v2} model, creating edges between arguments when similarity exceeds 70\%. Keyword detection utilizes the KeyBERT model, considering unigram to trigram phrases with a minimum importance score threshold of 0.4. For entity detection, we leverage Spacy and \href{https://github.com/egerber/spaCy-entity-linker}{Spacy-entity-linker} for both recognition and linking of entities (e.g., mapping "NRA" to "National Rifles Association"). We implement community detection using the Eva algorithm from Python's Cdlib library with default parameters. For all graph-related operations, including the creation of Minimum Spanning Trees, we utilize Python's NetworkX library. 
\laks{How is the choice of the number of communities ensuring fairness?}
\rudraorange{Please, see the paragraph below.}
\rudrabluetext{The number of communities generated by the baselines and our approaches (\ourapproachroberta{} and \ourapproachmst{}) can vary, potentially biasing evaluation metrics. To ensure a fair comparison with baselines, we standardized the evaluation by analyzing up to 6 communities for the Allsides dataset, 4 for the Perspectrum dataset, and 10 arguments per community for both. We applied Quality-based Filtering (Section~\ref{Budget-Constrained Optimization}) to select communities and arguments for the baselines. Due to the LLM token budget in \baselineiii{}, a maximum of 450 arguments could be included in the prompt. However, our approach does not rely on an external LLM for community generation, avoiding this limitation. We use \ourapproachbest{} to refer to the best-performing variant (STIC-[D/R]) throughout the remainder of this paper.}
\par
For perspective label generation, we utilize OpenAI's `GPT-3.5-turbo-instruct' model with temperature=0, max-tokens=200, top-p=1, and both frequency and presence penalties set to 1. The few-shot prompt template used for this task is provided in Section \ref{appendix:perspective-label-prompt}. The hyper-parameters are chosen using a combination of manual and experimental evaluation. For our approach, we use two NVIDIA P100 Pascal GPUs (16 GB memory each). The runtime for different topics ranges from fifteen minutes to one hour, depending on server load and GPU availability due to concurrent usage by other users on Compute Canada's infrastructure. For \baselineii{}, we utilize OpenAI's `gpt-3.5-turbo-16k' model with default parameters and temperature=0.

\begin{table*}[t]
\centering
\small
\begin{tabular}{l@{\hspace{1pt}}l c c c | l@{\hspace{1pt}}l c c c}
\toprule
Topic & Method & Coh & Info & Rel & Topic & Method & Coh & Info & Rel \\
\midrule
\multirow{5}{*}{Politics} & \baselinei{} & 1.15 & 1.46 & 1.18 & \multirow{5}{*}{Healthcare} & \baselinei{} & 1.37 & 1.57 & 1.62 \\
 & \baselineii{} & 2.16 & 2.37 & 2.32 & & \baselineii{} & 3.66 & \textbf{4.08} & \underline{4.52} \\
 & \baselineiii{} & 2.03 & 2.85 & 2.48 & & \baselineiii{} & 3.48 & 3.61 & 4.29 \\
 & \ourapproachroberta{} & $\underline{3.2}^3$ & $\textbf{3.52}^3$ & $\textbf{3.64}^3$ & & \ourapproachroberta{} & $\underline{3.68}^2$ & $3.96^3$ & $4.06^3$ \\
 & \ourapproachmst{} & $\textbf{3.46}^3$ & $\underline{3.35}^3$ & $\underline{3.47}^3$ & & \ourapproachmst{} & $\textbf{4.08}^3$ & $\underline{3.96}^3$ & $\textbf{4.56}^2$ \\
\midrule
\multirow{5}{*}{Elections} & \baselinei{} & 1.49 & 1.84 & 1.63 & \multirow{5}{*}{Terrorism} & \baselinei{} & 1.33 & 1.64 & 1.77 \\
 & \baselineii{} & 2.27 & \underline{3.51} & 2.99 & & \baselineii{} & \underline{3.48} & 3.91 & 4.00 \\
 & \baselineiii{} & 2.38 & 3.41 & 3.07 & & \baselineiii{} & 3.39 & \underline{4.09} & \textbf{4.52} \\
 & \ourapproachroberta{} & $\textbf{4.03}^3$ & $\textbf{3.96}^3$ & $\textbf{4.34}^3$ & & \ourapproachroberta{} & $\textbf{4.35}^3$ & $\textbf{4.12}^3$ & $4.28^2$ \\
 & \ourapproachmst{} & $\underline{2.84}^3$ & $3.12^3$ & $\underline{3.86}^3$ & & \ourapproachmst{} & $3.36^2$ & $3.54^3$ & $\underline{4.32}^2$ \\
\midrule
\multirow{5}{*}{\makecell[l]{Gun \\Control}} & \baselinei{} & 1.93 & 1.92 & 2.17 & \multirow{5}{*}{Education} & \baselinei{} & 1.70 & 2.11 & 2.27 \\
 & \baselineii{} & 3.7 & 4.21 & 4.64 & & \baselineii{} & 2.67 & 3.73 & \underline{3.68} \\
 & \baselineiii{} & \underline{4.3} & \underline{4.33} & \underline{4.87} & & \baselineiii{} & 0 & 0 & 0 \\
 & \ourapproachroberta{} & $\textbf{4.56}^3$ & $\textbf{4.58}^3$ & $\textbf{4.97}^3$ & & \ourapproachroberta{} & $\underline{3.49}^3$ & $\underline{3.78}^2$ & $3.63^2$ \\
 & \ourapproachmst{} & $3.9^3$ & $4.11^3$ & $4.74^1$ & & \ourapproachmst{} & $\textbf{3.92}^3$ & $\textbf{4.04}^3$ & $\textbf{3.85}^2$ \\
\midrule
\multirow{5}{*}{\makecell[l]{Immi-\\gration}} & \baselinei{} & 1.45 & 1.67 & 2.05 & \multirow{5}{*}{\makecell[l]{Supreme\\Court}} & \baselinei{} & 1.59 & 2.02 & 2.44 \\
 & \baselineii{} & 3.69 & 4.01 & 4.5 & & \baselineii{} & 3.54 & 3.79 & 4.25 \\
 & \baselineiii{} & 0 & 0 & 0 & & \baselineiii{} & 3.63 & \underline{4.13} & \underline{4.71} \\
 & \ourapproachroberta{} & $\underline{3.86}^3$ & $\textbf{4.12}^2$ & $\textbf{4.58}^2$ & & \ourapproachroberta{} & $\textbf{4.00}^3$ & $\textbf{4.27}^3$ & $\textbf{4.81}^3$ \\
 & \ourapproachmst{} & $\textbf{3.92}^3$ & $\underline{4.07}^2$ & $\underline{4.52}^2$ & & \ourapproachmst{} & $\underline{3.65}^1$ & $3.99^3$ & $4.45^3$ \\
\midrule
\multirow{5}{*}{\makecell[l]{Foreign\\Policy}} & \baselinei{} & 1.48 & 1.84 & 2.02 & \multirow{5}{*}{\makecell[l]{National\\Security}} & \baselinei{} & 1.44 & 1.51 & 1.49 \\
 & \baselineii{} & 2.63 & 3.51 & \underline{3.77} & & \baselineii{} & 2.95 & 3.51 & 3.51 \\
 & \baselineiii{} & 2.37 & \underline{3.55} & 3.73 & & \baselineiii{} & 0 & 0 & 0 \\
 & \ourapproachroberta{} & $\textbf{3.59}^3$ & $\textbf{4.16}^3$ & $\textbf{4.02}^3$ & & \ourapproachroberta{} & $\textbf{3.86}^3$ & $\textbf{3.88}^3$ & $\textbf{4.07}^3$ \\
 & \ourapproachmst{} & $\underline{2.85}^3$ & $3.09^3$ & $3.76^1$ & & \ourapproachmst{} & $\underline{3.43}^3$ & $\underline{3.47}^2$ & $\underline{3.54}^2$ \\
\bottomrule
\end{tabular}
\caption{Comparison of our approaches against the baseline models for ten topics from the Allsides dataset (Politics, Elections, Gun control, Immigration, Foreign Policy, Healthcare, Terrorism, Education, Supreme Court and National Security).}
\label{tab:allsides-performance}
\end{table*}

\subsection{Quantitative Analysis}\label{quantitative-analysis}
Our unsupervised approach organizes arguments across a large collection of articles in a complex, inter-article setting. \LL{To our knowledge,} no benchmark datasets exist for this level of argument organization, making human evaluation the ideal standard for the evaluation. However, human studies are resource-intensive for evaluating a large number of argument clusters. To address this, we adopted an automated evaluation inspired by \citet{liu2303gpteval}, which shows GPT-4’s alignment with human judgments in tasks like summarization. We selected three metrics—informativeness, relevance, and coherence—to ensure our system delivers clear, relevant, and well-structured argument communities for users. These metrics, widely used in prior work \citep{fabbri2021summeval, kryscinski2019evaluating}, enable fair comparisons with other approaches. Using GPT-4, we evaluated these metrics on a 1–5 scale (5 being the highest) across 10 zero-shot trials, reporting average scores for each system’s communities (details in Section \ref{appendix:prompts-for-evaluation}).


Experimental results on both datasets (Tables \ref{tab:allsides-performance}, \ref{tab:allsides-percentile-performance}, and \ref{tab:perspectrum-performance}) demonstrate the effectiveness of our approach. The best performance for each metric is \textbf{bolded}, while the second-best is \underline{underlined}. Superscripts (3, 2, 1) next to the values show the strength of statistical evidence from pairwise T-Tests, comparing our method to three baselines. \LL{A superscript of $i$ indicates strong evidence (p < 0.05) that our method outperforms baseline $i$, $i\in\{1,2,3\}$.} Higher numbers reflect stronger evidence of improvement. Our methods consistently achieve better results than the baselines across all topics and metrics, with these improvements confirmed to be statistically significant.


On Allsides, our approach achieves superior performance across \textbf{8} out of \textbf{10} topics. Specifically: \ourapproachmst{} excels in  Healthcare and Education, achieving coherence scores of \textbf{4.08} and \textbf{3.92}, respectively, and relevance scores of \textbf{4.56} and \textbf{3.85}. 
\ourapproachroberta{} outperforms the baselines in the remaining domains, particularly in Elections, and Gun control and gun rights, where it achieves coherence scores of \textbf{4.03}, and \textbf{4.56}, respectively, and relevance scores of \textbf{4.34} and \textbf{4.97}. On Perspectrum, our approach demonstrates superior performance across \textbf{all 4 topics}, with statistically significant improvements in coherence, informativeness, and relevance. For instance, in the topic ``Abolish nuclear weapons'', \ourapproachroberta{} achieves coherence, informativeness, and relevance scores of \textbf{4.30}, \textbf{4.39}, and \textbf{4.66}, respectively, outperforming all baselines. This demonstrates its robustness in handling diverse and complex argument structures while maintaining high coherence and relevance.

\begin{table*}[t]
\centering
\small
\begin{tabular}{l l c c c}
\toprule
Topic & Method & Coh & Info & Rel \\
\midrule
\multirow{5}{*}{\makecell[l]{Abolish \\ nuclear \\ weapons}} 
& \baselinei{} & 1.84 & 1.89 & 2.29 \\
& \baselineii{} & 3.62 & 4.20 & 4.50 \\
 & \baselineiii{} & 3.35 & 3.98 & 4.24 \\
 & \ourapproachroberta{} & $\underline{4.30}^3$ & $\textbf{4.39}^3$ & $\textbf{4.66}^3$ \\
 & \ourapproachmst{} & $\textbf{4.41}^3$ & $\underline{4.23}^2$ & $\underline{4.24}^2$ \\
\midrule
\multirow{5}{*}{\makecell[l]{All nations \\ have a right to \\ nuclear \\ weapons}}
& \baselinei{} & 1.71 & 1.93 & 1.98 \\
& \baselineii{} & 3.98 & 4.06 & 4.62 \\
 & \baselineiii{} & 4.13 & \underline{4.28} & 4.43 \\
 & \ourapproachroberta{} & $\textbf{4.52}^3$ & $\textbf{4.35}^3$ & $\underline{4.66}^2$ \\
 & \ourapproachmst{} & $\underline{4.21}^2$ & $3.94^2$ & $\textbf{4.71}^3$ \\
\midrule
\multirow{4}{*}{\makecell[l]{Social net-\\ working sites \\ are good for \\ our society}} 
& \baselinei{} & 1.60 & 1.79 & 1.62 \\
& \baselineii{} & 3.58 & 4.12 & 4.14 \\
 & \baselineiii{} & \underline{4.01} & \textbf{4.26} & 3.84 \\
 & \ourapproachroberta{} & $\textbf{4.19}^2$ & $\underline{4.17}^1$ & $\textbf{4.44}^3$ \\
 & \ourapproachmst{} & $3.36^2$ & $3.79^3$ & $\underline{4.21}^2$ \\
\midrule
\multirow{4}{*}{\makecell[l]{There is a need \\ for developing \\ tactical nuclear \\ weapons}}
& \baselinei{} & 1.68 & 1.77 & 1.68 \\
& \baselineii{} & 3.87 & 3.99 & \underline{4.46} \\
 & \baselineiii{} & 3.22 & 4.08 & 4.18 \\
 & \ourapproachroberta{} & $\underline{4.43}^3$ & $\textbf{4.36}^3$ & $\textbf{4.69}^3$ \\
 & \ourapproachmst{} & $\textbf{4.50}^3$ & $\underline{4.22}^3$ & $4.10^2$ \\
\bottomrule
\end{tabular}
\caption{Comparison of our approaches against the baseline models for 4 topics on the \textbf{Perspectrum} Dataset.}
\label{tab:perspectrum-performance}
\end{table*}

\subsection{Baseline Comparisons}\label{baseline-comparisons}

\baselinei{}, which is based solely on clustering of argument embedding, consistently underperforms across all metrics and topics. For example, in the Immigration topic, it achieves coherence and relevance scores of only \textbf{1.45} and \textbf{2.05}, respectively, highlighting the limitations of embedding-only approaches. \baselineiii{}, despite employing ChatGPT-like models for key point generation and matching, demonstrates significant parsing challenges, particularly in the Immigration and Education topics, where it fails to produce meaningful output (rows with \textbf{0} performance in Table \ref{tab:allsides-performance}). \baselineii{} performs better than \baselinei{} and \baselineiii{} in some domains, such as Healthcare and Supreme Court, but still falls short of \ourapproachbest{}. For instance, in the National Security topic, \baselineii{} achieves a relevance score of \textbf{3.51}, which is surpassed by \ourapproachmst{}'s score of \textbf{4.07}.

\subsection{Metric-wise Performance Analysis}\label{metric-wise-performance-analysis}

Performance analysis reveals nuanced insights into the challenges of argument organization. Coherence emerges as the most challenging metric, with baseline methods struggling to maintain semantic and structural cohesion. For example, \baselinei{} achieves coherence scores below 2.0 across most topics. Our approach significantly improves Informativeness and Relevance. For instance, \ourapproachroberta{} achieves the highest informativeness score of \textbf{4.58} in the Gun control topic, demonstrating its ability to present meaningful arguments.

\begin{table*}[t]
\centering
\small
\begin{tabular}{l@{\hspace{1pt}}l c c c | l@{\hspace{1pt}}l c c c}
\toprule
Topic & Method & Coh & Info & Rel & Topic & Method & Coh & Info & Rel \\
\midrule
\multirow{5}{*}{Politics} & AEC & 0.0 & 10.0 & 0.0 & \multirow{5}{*}{Healthcare} & AEC & 0.0 & 3.33 & 3.33 \\
 & KPA & 3.33 & 5.0 & 11.67 & & KPA & 55.0 & \textbf{91.67} & \textbf{100.0} \\
 & KPA-GPT & 3.33 & 13.33 & 18.33 & & KPA-GPT & 45.0 & 53.33 & 91.67 \\
 & STIC-R & 36.67 & 46.67 & \textbf{51.67} & & STIC-R & 66.0 & 74.0 & 74.0 \\
 & STIC-D & \textbf{50.0} & \textbf{48.33} & \textbf{51.67} & & STIC-D & \textbf{83.33} & 80.0 & 91.67 \\
\midrule
\multirow{5}{*}{Elections} & AEC & 1.67 & 18.33 & 1.67 & \multirow{5}{*}{Terrorism} & AEC & 0.0 & 11.67 & 15.0 \\
 & KPA & 1.67 & 55.0 & 30.0 & & KPA & 45.0 & 85.0 & 75.0 \\
 & KPA-GPT & 10.0 & 41.67 & 21.67 & & KPA-GPT & 38.0 & 88.0 & \textbf{92.0} \\
 & STIC-R & \textbf{72.0} & \textbf{80.0} & \textbf{92.0} & & STIC-R & \textbf{86.67} & \textbf{90.0} & 88.33 \\
 & STIC-D & 18.33 & 23.33 & 78.33 & & STIC-D & 36.67 & 48.33 & 90.0 \\
\midrule
\multirow{5}{*}{\makecell[l]{Gun \\Control}} & AEC & 28.33 & 18.33 & 18.33 & \multirow{5}{*}{Education} & AEC & 8.33 & 21.67 & 25.0 \\
 & KPA & 65.0 & 85.0 & 95.0 & & KPA & 8.33 & 65.0 & 70.0 \\
 & KPA-GPT & 93.33 & 95.0 & 100.0 & & KPA-GPT & 0.0 & 0.0 & 0.0 \\
 & STIC-R & \textbf{100.0} & \textbf{100.0} & \textbf{100.0} & & STIC-R & 53.33 & 68.33 & 63.33 \\
 & STIC-D & 76.67 & 86.67 & 96.67 & & STIC-D & \textbf{78.33} & \textbf{90.0} & \textbf{75.0} \\
\midrule
\multirow{5}{*}{\makecell[l]{Immi-\\gration}} & AEC & 8.33 & 16.67 & 26.67 & \multirow{5}{*}{\makecell[l]{Supreme\\Court}} & AEC & 5.0 & 16.67 & 23.33 \\
 & KPA & 63.33 & 81.67 & 80.0 & & KPA & 53.33 & 78.33 & 83.33 \\
 & KPA-GPT & 0.0 & 0.0 & 0.0 & & KPA-GPT & 55.0 & 95.0 & 95.0 \\
 & STIC-R & 56.67 & 83.33 & 91.67 & & STIC-R & \textbf{83.33} & \textbf{96.67} & \textbf{100.0} \\
 & STIC-D & \textbf{73.33} & \textbf{86.67} & \textbf{91.67} & & STIC-D & 61.67 & 86.67 & 98.33 \\
\midrule
\multirow{5}{*}{\makecell[l]{Foreign\\Policy}} & AEC & 0.0 & 15.0 & 21.67 & \multirow{5}{*}{\makecell[l]{National\\Security}} & AEC & 5.0 & 13.33 & 13.33 \\
 & KPA & 15.0 & 43.33 & 63.33 & & KPA & 23.33 & 61.67 & 58.33 \\
 & KPA-GPT & 1.67 & 55.0 & 58.33 & & KPA-GPT & 0.0 & 0.0 & 0.0 \\
 & STIC-R & \textbf{50.0} & \textbf{88.33} & \textbf{81.67} & & STIC-R & \textbf{68.33} & \textbf{80.0} & \textbf{81.67} \\
 & STIC-D & 15.0 & 30.0 & 63.33 & & STIC-D & 48.33 & 48.33 & 48.33 \\
\bottomrule
\end{tabular}
\caption{Performance comparison between our approaches and baseline models across ten topics from the Allsides dataset. Each cell shows the percentage of trials (out of 60) in which the model received a score of at least 4 out of 5.}
\label{tab:allsides-percentile-performance}
\end{table*}

\rudraorange{I have added a subsection below to address one of the reviewer's questions, C19}
\subsection{\rudrabluetext{Detailed Analysis on the Allsides Dataset}}\label{detailed-analysis-allsides-dataset}

Table \ref{tab:allsides-percentile-performance} presents a comprehensive evaluation of GPT-4 scores for communities generated by our approaches (\ourapproachroberta{} and \ourapproachmst{}) and baseline methods (\baselinei{}, \baselineii{}, \baselineiii{}) across ten topics from the Allsides dataset. The percentage values indicate the frequency with which each method achieved a score of at least 4 out of 5 in coherence, informativeness, and relevance over 60 trials.
\par
Our approaches outperform the baselines in most topics \LL{across all three metrics}, with \ourapproachroberta{} and \ourapproachmst{} achieving the highest percentages for scores $\geq 4$ in 9 out of 10 topics. \LL{Two exceptions are Healthcare and Terrorism:  in Healthcare, \baselineii{} excels, whereas in Terrorism, \baselineiii{} achieves a superior performance on relevance.}  Specifically, \ourapproachroberta{} achieves high performance (scores $\geq 4$ at least 80\% of the time) in 17 out of 30 evaluation cases across the 10 topics and 3 metrics, demonstrating superior performance in generating coherent, informative, and relevant summaries. \laks{how can i verify the 80\% claim?} For example, STIC-R achieves 92.0\% performance in Elections (relevance) and perfect 100.0\% scores across all three metrics in Gun Control — representing 4 of the 17 high-performing cases where it significantly outperforms all baseline methods. In contrast, \baselineii{} and \baselineiii{} reach scores $\geq 4$ at least 80\% of the time in only 8 out of the 30 cases, with \baselineii{} performing strongly in Healthcare (91.67\% informativeness, 100.0\% relevance) and Gun Control, but trailing elsewhere.
\par
\ourapproachmst{} also shows competitive performance, leading in Politics (50.0\% coherence), Healthcare (83.33\% coherence), Education (78.33\% coherence, 90.0\% informativeness), and Immigration (73.33\% coherence, 86.67\% informativeness). However, its performance is less consistent than \ourapproachroberta{}, particularly in Elections (18.33\% coherence) and Foreign Policy (15.0\% coherence), suggesting that its Minimum Spanning Tree simplification may occasionally disrupt argumentative flow. The \baselinei{} baseline consistently underperforms, with near-zero percentages in Politics and Healthcare coherence, indicating its inability to form cohesive communities.

\rudraorange{I have added a subsection below to address one of the reviewer's questions, C19}
\subsection{\rudrabluetext{Case Study on Community Quality}}

\LL{We report on a case study on community quality that we conducted. For this purpose, we chose the topics `Elections' and `Education' from the Allsides dataset. 
Figures \ref{fig:KPA-sample-community-output} and \ref{fig:KPA-GPT-sample-community-output} illustrate two communities 
generated by \baselineii{} and \baselineiii{}, respectively, which GPT-4 rated poorly (less than or equal to 2.5 on a 1--5 scale) in coherence, informativeness, and relevance.} In Figure \ref{fig:KPA-sample-community-output}, the arguments lack coherence, switching context between unrelated events and figures, such as Canadian elections, Hillary Clinton, and Romney’s EPA attacks (arguments 1, 2, and 9), without logical connections. This fragmentation obscures the central argument of `Election', focusing on personal attacks (arguments 3 and 7), which dilute relevance. Similarly, Figure \ref{fig:KPA-GPT-sample-community-output} presents fragmented arguments covering disparate topics, from an assault-weapons ban to Romney’s foreign policy and cultural disputes, with no clear transitions or electoral focus. For instance, the ’94 crime bill  (argument 1) and cultural conflicts (argument 9) have no significant connection to elections, while points like Ohio’s role in politics (argument 8) are too generic to offer insight. 
\par
In contrast, Figure \ref{fig:STIC-R-sample-community-output-good-quality} showcases a community from \ourapproachroberta{} on `Elections,' rated highly by GPT-4 (greater than or equal to 4 on a 1--5 scale) on all metrics. The arguments form a logically connected hierarchy, as reflected in the DFS numbering. The top-level point about Trump’s post-debate polling (argument 1) sets the context, with sub-points exploring related themes, such as debate significance  (argument 1.1), qualification rules (argument 1.1.1), and Biden’s candidacy (argument 1.2). Each level builds on the previous, ensuring a clear, nested structure that enhances readability. The cluster provides specific, fact-based insights, citing sources like a Quinnipiac University survey (argument 1.1.1.1.1), avoiding vague or opinionated claims. All arguments directly tie to elections, focusing on debates, polling, and campaign strategies, with the DFS structure maintaining relevance throughout.
\par
However, Figure \ref{fig:STIC-R-sample-community-output-poor-quality} presents a community from \ourapproachroberta{} on `Education,' rated poorly by GPT-4 (less than or equal to 2.5 on a 1--5 scale). Despite the DFS hierarchy, the arguments lack logical flow. The top-level point about school shooters and discriminatory practices (argument 1) is unrelated to sub-branches discussing textbook inaccuracies (argument 1.1), labor unions (argument 1.1.1), and meritocracy. Another branch on cognitive skills (argument 1.2) and educational reform fails to connect to the root argument, resulting in a fragmented cluster and limiting its informativeness. 
Further analysis of such poorly rated communities reveals that the interaction graph’s edge types (topic, semantic, keyword, entity) can sometimes produce weak or superficial connections. For instance, arguments connected by a shared entity, such as a person, organization, or a broad generic topic, may differ substantially in context, leading to incoherent communities. As illustrated in Figure~\ref{fig:STIC-R-sample-community-output-poor-quality}, most arguments are linked primarily through the generic topic ``education,'' which offers little thematic cohesion. In contrast, Figure~\ref{fig:STIC-R-sample-community-output-good-quality} shows arguments connected by highly specific and contextually relevant keywords such as ``presidential debate,'' ``Republican,'' and ``Donald Trump,'' which are directly pertinent to the election topic. Moreover, degree- and quality-based filtering often fails to eliminate such loosely related or irrelevant arguments, further reducing the overall coherence and relevance of the community.


\begin{figure}[!htb]
  \centering
  \begin{subfigure}{\textwidth}
    \centering
    \begin{tikzpicture}
      \draw[thick] (0,0) rectangle (13, -7.8);
      \node[anchor=north west, align=left, text width=12.3cm] at (0.2, -0.5) {
        \textbf{Community 5} \\
        \vspace{0.2cm}
        \textbf{Perspective Label}: 2019 Election Campaign \\
        \vspace{0.2cm}
        1. The 2019 election campaign was one of the most divisive and negative in Canadian history. \newline
        2. We know that Hillary Clinton is the worst, most horrifying and certainly lethal poison known to man. \newline
        3. It was an astonishing admission: After an episode that wrecked his political career, nearly destroyed his marriage, and made him an object of international mockery. \newline
        ......... \newline
        7. Bush attacked Trump for a lack of seriousness, noting snidely at one point that he was giving him a bit of your own medicine. \newline
        8. They catcall the media and shove and hit protesters while Trump cheers them on from the stage. \newline
        9. Romney intensifies EPA attacksMitt. \newline
        10. Fellow GOP pols don’t like him. \newline
      };
    \end{tikzpicture}
    \caption{An abbreviated example of a poorly rated community on the topic `Elections', generated by \textbf{\baselineii{}}}
    \label{fig:KPA-sample-community-output}
  \end{subfigure}

  \vspace{0.3cm} 

  \begin{subfigure}{\textwidth}
    \centering
    \begin{tikzpicture}
      \draw[thick] (0,0) rectangle (13, -8.2);
      \node[anchor=north west, align=left, text width=12.3cm] at (0.2, -0.5) {
        \textbf{Community 3} \\
        \vspace{0.2cm}
        \textbf{Perspective Label}: Partisan Conflict and Cultural Division in Electoral Politics \\
        \vspace{0.2cm}
        1. An assault-weapons ban was included in the ’94 crime bill along with other legislation .... a pattern or practice of local law enforcement abuses. \newline
        2. Mitt Romney in a major speech Monday will call for a change of course in U.S. foreign policy ... hope is not a strategy. \newline
        3. All it took was the adolescent agitprop of Donald Trump to lay bare the rift  ...  whose anxieties and prejudices they have been manipulating for decades. \newline
        4. Biden seemed more vulnerable than ever coming into the debate, facing persistent questions about his age, verbal slip-ups and performance on the campaign trail. \newline
        ...... \newline
        8. Ohio plays a crucial role in campaign politics. \newline
        9. The left demanded cultural homogeneity and is outraged that Americans refuse to go along with their anti-Christian secular agenda. \newline
        10. Conservatives and people of faith have been hounded from their jobs, silenced, and punished for refusing to go along with the left’s agenda. \newline
      };
    \end{tikzpicture}
    \caption{An abbreviated example of a poorly rated community on the topic `Elections', generated by \textbf{\baselineiii{}}.}
    \label{fig:KPA-GPT-sample-community-output}
  \end{subfigure}
  \caption{Abbreviated examples of poorly rated communities on the topic `Elections', generated by \textbf{\baselineii{}} and \textbf{\baselineiii{}} and evaluated by GPT-4.}
  \label{fig:combined-KPA-sample-community-output}
\end{figure}

\begin{figure}[p]
  \centering
  \begin{subfigure}{\textwidth}
    \centering
    \begin{tikzpicture}
      \draw[thick] (0,0) rectangle (13, -8.7);
      \node[anchor=north west, align=left, text width=12.3cm] at (0.2, -0.5) {
        \textbf{Community 1} \\
        \vspace{0.2cm}
        \textbf{Perspective Label}: Legitimacy and Power Struggles in U.S. Presidential Elections  \\
        \vspace{0.2cm}
        1. Polls conducted since the first presidential debate last month put Donald Trump on a pace to earn a smaller percentage of the vote than any major-party nominee in at least 20 years. \newline
        1.1. The first presidential debate tonight is shaping up to be one of the most-watched political events ever, with a potentially Super Bowl-size audience. \newline
        1.1.1. Democratic National Committee imposed tougher rules for presidential candidates to qualify for primary debates scheduled for September and October, .... \newline
        1.1.1.1. A beleaguered Donald Trump sought to undermine the legitimacy of the U.S. presidential election on Saturday, .... baseless insinuation his rival was on drugs in the last debate. \newline
        1.1.1.1.1. According to a Quinnipiac University survey of Ohio voters, Clinton remains in the lead over eight potential Republican opponents ... but her margins have shrunk. \newline
        ..... \newline
        1.2. Joe Biden officially announced his presidential bid this morning, promising that he could save the country from Donald Trump and the forces of white supremacy. \newline
      };
    \end{tikzpicture}
    \caption{An abbreviated example of a highly rated community on the topic `Elections', generated by \textbf{\ourapproachroberta{}}.}
    \label{fig:STIC-R-sample-community-output-good-quality}
  \end{subfigure}

  \vspace{0.2cm} 

  \begin{subfigure}{\textwidth}
    \centering
    \begin{tikzpicture}
      \draw[thick] (0,0) rectangle (13, -8.7);
      \node[anchor=north west, align=left, text width=12.3cm] at (0.2, -0.5) {
        \textbf{Community 1} \\
        \vspace{0.2cm}
        \textbf{Perspective Label}: Critique of Structural Inequities and Meritocracy in Education \\
        \vspace{0.2cm}
        1. One point that critics have brought up is that school shooters, unlike the minority students ... of discriminatory disciplinary practices, are usually white. \newline
        1.1. But, a popular AP U. S. History textbook, The American Pageant, written by Thomas Bailey, David Kennedy, and Lizabeth Cohen, is also riddled with inaccuracies. \newline
        1.1.1. SEIU is the only labor organization in the U. S. that has been even more politically active in recent decades than the teachers’ unions. \newline
        ..... \newline
        1.1.1.1.1.1. The move has national implications, since textbooks retooled to fit California’s changing history frameworks are often used much more widely. \newline
        ...... \newline
        1.2. They struggle with questions that require higher cognitive demands, such as taking a real-world situation and translating it into a formula, according to the report. \newline
        1.2.1. It’s a seismic shift in education meant to better prepare kids for college, career and the global economy. \newline
      };
    \end{tikzpicture}
    \caption{An abbreviated example of a poorly rated community on the topic `Education', generated by \textbf{\ourapproachroberta{}}.}
    \label{fig:STIC-R-sample-community-output-poor-quality}
  \end{subfigure}
  \caption{Examples of communities on the topics `Elections' (highly rated) and `Education' (poorly rated), generated by \textbf{\ourapproachroberta{}} and evaluated by GPT-4.}
  \label{fig:combined-STIC-R-sample-community-output}
\end{figure}

\subsection{Qualitative Analysis} \label{ref:qualitative-Analysis}

We conducted a human evaluation study to assess outputs of two baseline systems\footnote{\baselinei{} was excluded due to consistently poor performance in quantitative analysis.} and our best approach (STIC*, STIC-D or STIC-R, based on quantitative performance)  across five topics: Gun Control, Terrorism, Supreme Court, Elections, and Healthcare. 
The study began with a qualification test, where 70 Amazon Mechanical Turk workers answered seven questions on system outputs via Qualtrics. The study was conducted in two phases. First, we administered a qualification test using Amazon Mechanical Turk and Qualtrics, inviting 70 workers (referred to as "Turkers") to evaluate outputs from the three systems. The qualification test consisted of seven questions to be completed within one hour. To ensure high-quality annotations, we selected Turkers with the following criteria: (1) more than 10,000 HITs approved, (2) a HIT approval rate greater than 95\% and (3) location restricted to the United States or Canada. Among participants, the top 15 performers who achieved scores of $\geq 5/7$ were invited to complete the follow-up survey. Nine participants completed the survey following the provided guidelines. Participants were compensated at a rate of \$15 per hour for both the qualification test and survey completion. The top 15 performers (scoring $\geq 5/7$) were invited for the survey (more details on survey instruction can be found in Section \ref{appendix:survey-instruction}). Turkers scored each system (on a 1-5 scale, 5 being the highest) on four predefined metrics: a) \textbf{Diversity of Viewpoints}: How useful are the communities in a system in bringing out a diversity of viewpoints as well as stances? b) \textbf{Informativeness}: How well do the arguments in a community collectively provide comprehensive information about the given topic and its associated viewpoints? c) \textbf{Coherence}: How well do the arguments in a community flow and connect to form a meaningful discussion of the topic? d) \textbf{Overall Quality}: Overall, how do you rate the system concerning the way arguments are organized? Snippets of the outputs from \baselineii{}, \baselineiii{} and \ourapproachroberta{} are shown in Figures \ref{fig:baselineii-survey-output}, \ref{fig:baselineiii-survey-output} and \ref{fig:stic-r-survey-output-first-part} respectively.

\begin{table*}[t]
\centering
\small
\begin{tabular}{llcccc}
\toprule
Topic & Method & \makecell{Diversity of \\ Viewpoints} & Info & Coh & \makecell{Overall \\ Quality} \\
\midrule
\multirow{3}{*}{Election} & \baselineii{} & 2 & 3 & 2 & 2 \\
 & \baselineiii{} & 1 & 3 & 2 & 2 \\
 & \ourapproachbest{} & \textbf{4} & \textbf{4} & \textbf{4} & \textbf{4} \\
\midrule
\multirow{3}{*}{\makecell[l]{Gun \\ Control}} & \baselineii{} & 3 & 4 & \textbf{4} & 3 \\
 & \baselineiii{} & 2 & 4.5 & 3 & \textbf{4.5} \\
 & \ourapproachbest{} & \textbf{4.5} & \textbf{4.5} & 3.5 & 4 \\
\midrule
\multirow{3}{*}{Healthcare} & \baselineii{} & 2 & 3 & 3 & 3 \\
 & \baselineiii{} & 3 & 4 & 4 & 4 \\
 & \ourapproachbest{} & \textbf{3} & \textbf{4} & \textbf{5} & \textbf{4} \\
\midrule
\multirow{3}{*}{\makecell[l]{Supreme \\ Court}} & \baselineii{} & 4 & 4.33 & 3.33 & 2.67 \\
 & \baselineiii{} & 3.67 & \textbf{4.33} & \textbf{4} & 4 \\
 & \ourapproachbest{} & \textbf{4.67} & 3.67 & 3.67 & \textbf{4} \\
\midrule
\multirow{3}{*}{Terrorism} & \baselineii{} & 4 & 2.5 & 4 & 4 \\
 & \baselineiii{} & 4 & 2.5 & \textbf{4.5} & \textbf{4.5} \\
 & \ourapproachbest{} & \textbf{4} & \textbf{2.5} & 4 & 4 \\
\bottomrule
\end{tabular}
\caption{Qualitative analysis results on the \textbf{Allsides} dataset.}
\label{tab:survey-results}
\end{table*}

\subsection{Human Study Results}\label{human-study-results}

\laks{"our approach" gets pretty ambiguous here, given that you don't say what STIC* is: is it both or at least one of R and D? also, why present results in this way all of a sudden?}
Table \ref{tab:survey-results} presents the human study results, highlighting key findings: STIC* achieved the highest scores in \textbf{Diversity of Viewpoints} across \textbf{four} topics (Elections: 4, Gun Control: 4.5, Healthcare: 3, Supreme Court: 4.67) and tied for best in Terrorism (4), demonstrating its ability to reveal a spectrum of viewpoints. In \textbf{Informativeness}, STIC* outperformed baselines in \textbf{three} topics (Elections: 4, Gun Control: 4.5, Healthcare: 4) and tied for best in Terrorism (2.5), aligning with quantitative results. For \textbf{Coherence}, STIC* was rated best in two topics (Elections: 4, Healthcare: 5), while \baselineiii{} performed better in Gun Control (4.5) and Terrorism (4.5), with Supreme Court showing competitive performance across methods. This suggests that while STIC* excels in maintaining coherence in certain domains, there is room for improvement in others. In \textbf{Overall Quality}, STIC* was rated best or tied for best in \textbf{three} topics (Elections: 4, Healthcare: 4, Supreme Court: 4), with \baselineiii{} performing better in Gun Control (4.5) and Terrorism (4.5).



\subsection{Comparison between Baselines for Human Study} \label{comparison-between-baselines}

The baseline comparisons yield critical observations:
\begin{itemize}
    \item \textbf{\baselineii{}}: This baseline performs moderately well in some topics, such as \textbf{Gun Control} and \textbf{Supreme Court}, but consistently underperforms in others, such as \textbf{Elections} and \textbf{Healthcare}. Its limitations are particularly evident in the diversity of viewpoints, where it achieves scores of 2 or 3 across most topics.
    \item \textbf{\baselineiii{}}: This baseline performs better than \baselineii{} in several domains, particularly in coherence and overall quality. For example, in \textbf{Gun Control}, it achieves a coherence score of 4 and an overall quality score of 4.5. However, it struggles with diversity of viewpoints, achieving scores of 1 or 2 in \textbf{Elections} and \textbf{Gun Control}.
\end{itemize}

\subsection{Comparison Between Quantitative and Qualitative Results}\label{comparison-between-quantitative-and-qualitative-results}
The human study results largely align with quantitative findings but show some discrepancies. In \textbf{Healthcare} and \textbf{Elections}, both evaluations agree that our approach outperforms baselines in coherence, informativeness, and overall quality. For example, in \textbf{Healthcare}, our approach achieves a coherence score of 5 in the human study, matching its strong quantitative performance (Table \ref{tab:allsides-performance}). There are some exceptions: in \textbf{Gun Control} and \textbf{Terrorism}, the human study shows \baselineiii{} outperforming our approach in coherence and overall quality. 
However, we stress that, unlike the human study (Table \ref{tab:survey-results}), the quantitative findings (Table \ref{tab:allsides-performance}) come with statistical significance (with \(p < 0.05\)).
Our analysis revealed that \ourapproachroberta{} outperformed most baselines, hypothesized to two key factors: (i) more informative arguments generated through Stance Tree construction, and (ii) more cohesive communities resulting from the interaction graph's design. However, in cases where our system underperformed compared to baselines, the primary issue was imperfect argument pruning by the RoBERTa model. Despite training, the model occasionally fails to filter out irrelevant arguments, which reduces the overall quality of a community.

%% file: conclusion.tex
\section{Conclusion and Future Work}

We introduce a novel community-based approach to argument organization by leveraging diverse semantic relationships \LL{between arguments} to capture the complex interplay of viewpoints on contentious topics. Empirical results show that our method identifies nuanced perspectives often missed by traditional approaches and generates more informative summaries. One of our key contributions is our method's ability to systematically expose users to diverse, well-organized viewpoints, fostering informed discourse. By enabling efficient navigation of complex argumentative landscapes without manual processing, our approach makes a significant advancement in argument mining and organization. Future research directions include: (1) exploring the temporal dimension to analyze how viewpoints over controversial topics evolve over time, and (2) developing personalized argument navigation systems that balance viewpoint diversity with user interests.

%% file: limitations.tex
\section{Limitations}

While our approach demonstrates significant advancements in community-based argument organization, several limitations remain:

\squishlist
    
    \item \textbf{Human Evaluation Scope}: The human study, while insightful, was conducted on a limited set of topics i.e. Gun Control, Terrorism, Supreme Court, Elections, Healthcare and involved a relatively small group of turkers. A larger-scale study with more topics and participants could provide stronger evidence of our method's effectiveness.
        
    \item \textbf{Computational Cost}: The construction of interaction graphs and community detection, especially for large datasets, can be computationally expensive. Optimizing the scalability of our approach for real-time applications remains an exciting direction for future work.
    
    \item \textbf{Stance Simplification}: Our assumption that all arguments in an article share the same stance may oversimplify real-world scenarios where arguments within a single article can have varying stances. A more granular stance detection mechanism could improve accuracy.
\squishend

%% file: appendix.tex
\appendixsection{Prompts for Perspective Label Generation} \label{appendix:perspective-label-prompt}
Here are the manually created 4-shot prompts and instruction, used for perspective label generation:
\par
Few Shot Example 1: [TOPIC]: Vaccines. [ARGUMENTS]: \# Vaccines can save children’s lives. \# The American Academy of Pediatrics states that “most childhood vaccines are 90\%-99\% effective in preventing disease.” \# Vaccines can cause serious and sometimes fatal side effects. \# According to the CDC, all vaccines carry a risk of a life-threatening allergic reaction (anaphylaxis) in about one per million children. [OUTPUT]: Impact of vaccine on children's lives
\par
Few Shot Example 2: "[TOPIC]: Climate change. [ARGUMENTS]: \# Overwhelming scientific consensus finds human activity primarily responsible for climate change. \# Many scientists disagree that human activity is primarily responsible for global climate change. \# A paper published in Asia-Pacific Journal of Atmospheric Sciences found that some climate models overstated how much warming would occur from additional C02 emissions. \# According to a study published in the Journal of Atmospheric and Solar-Terrestrial Physics, 50-70\% of warming throughout the 20th century could be associated with an increased amount of solar activity. [Output]: Scientific studies regarding climate change
\par
Few Shot Example 3: [TOPIC]: Textbook vs Tablet. [ARGUMENTS]: \# Unlike tablets, there is no chance of getting malware, spyware, or having personal information stolen from a print textbook. \# On a tablet, e-textbooks can be updated instantly to get new editions or information. \# Tablets can hold hundreds of textbooks on one device, plus homework, quizzes, and other files, eliminating the need for physical storage of books and classroom materials. [Output]: Storage and security of textbook vs tablet
\par
Few Shot Example 4: [TOPIC]: College degrees. [ARGUMENTS]: \# Many college graduates are employed in jobs that do not require college degrees. \# College graduates have more and better employment opportunities. \# College graduates are more likely to have health insurance and retirement plans., Many recent college graduates are un- or underemployed. \# Many people succeed without college degrees. [Output]: Usefulness of college degrees
\par
Instruction: Given a topic and a list of arguments, create a phrase label that describes the perspective in the arguments. Each argument starts with a '\#'. The phrase label should act as a summarizer of the given text. Use the token '[Output]:' at the beginning of generation.

\appendixsection{Prompts for GPT-4 based Evaluation}\label{appendix:prompts-for-evaluation}

Here are the prompts we used for quantitative analysis. They largely follow the same pattern as shown in \href{https://github.com/nlpyang/geval/tree/main}{\cite{liu2303gpteval}'s work}:
\paragraph{\textbf{Coherence}} You will be given one community of arguments derived from political articles, along with a topic. Your task is to rate the argument community on one metric. Please make sure you read and understand these instructions carefully. Please keep this document open while reviewing, and refer to it as needed.\par

Evaluation Criteria:
Coherence (1-5) - how well arguments in the community flow and connect with each other to form a meaningful discussion of the topic. We align this dimension with how well the arguments build upon, contrast with, or relate to each other, regardless of their stance. The community should demonstrate logical connections between arguments rather than being a disconnected collection of statements about the topic.

Evaluation Steps:
\squishlist
\item Read the given topic and arguments to understand the context of discussion.
\item Read the argument community carefully and examine how the arguments relate to each other. Check if the arguments, regardless of their stance, form a coherent discussion where ideas connect, contrast, or build upon each other.
\item Assign a score for coherence on a scale of 1 to 5, where 1 is the lowest and 5 is the highest based on the Evaluation Criteria.
\squishend
Example:

Topic:

Argument Community:

Evaluation Form (scores ONLY):

- Coherence:

\paragraph{\textbf{Relevance}} You will be given one community of arguments derived from political articles, along with a topic. Your task is to rate the argument community on one metric. Please make sure you read and understand these instructions carefully. Please keep this document open while reviewing, and refer to it as needed.\par

Evaluation Criteria:
Relevance (1-5) - how well all arguments in the community align with and contribute to a specific viewpoint regarding the topic. We align this dimension with how well each argument connects to the central viewpoint being discussed. The community should not contain arguments that are tangential or unrelated to the viewpoint.

Evaluation Steps:
\squishlist
\item Read the argument community carefully and identify the central political viewpoint being represented.
\item Examine how well each argument relates to this viewpoint. Check if the arguments, regardless of their stance, are directly connected to the main viewpoint being discussed.
\item Assign a score for relevance on a scale of 1 to 5, where 1 is the lowest and 5 is the highest based on the Evaluation Criteria.
\squishend
Example:

Topic:

Argument Community:

Evaluation Form (scores ONLY):

- Relevance:

\paragraph{\textbf{Informativeness}} You will be given one community of arguments derived from political articles, along with a topic. Your task is to rate the argument community on one metric. Please make sure you read and understand these instructions carefully. Please keep this document open while reviewing, and refer to it as needed.\par

Evaluation Criteria:
Informativeness (1-5) - the collective quality of all arguments in the community. We align this dimension with how well the arguments collectively provide comprehensive and relevant information about the given topic and its associated viewpoints. The community should not just be a collection of loosely related arguments, but should present a coherent body of information that effectively communicates various perspectives on the topic.

Evaluation Steps:
\squishlist
\item Read the given topic and the argument community carefully.
\item Examine how well the arguments work together to provide meaningful information about the topic and its various viewpoints. Check if the arguments offer substantive support and development of different positions related to the topic.
\item Assign a score for informativeness on a scale of 1 to 5, where 1 is the lowest and 5 is the highest based on the Evaluation Criteria.
\squishend
Example:

Topic:

Argument Community:

Evaluation Form (scores ONLY):

- Informativeness:

\appendixsection{Human Study: Instruction}\label{appendix:survey-instruction}

Below, we provide the detailed instruction used for our human study.
 
\paragraph{\textbf{Background}} In discussions of controversial topics (like abortion or gun control), different arguments can represent left-leaning, center, or right-leaning perspectives. This survey examines how well different systems can organize related arguments into meaningful groups/communities. For optimal viewing experience, please use Google Chrome or Mozilla Firefox on a desktop or laptop computer. \par
 
\paragraph{\textbf{Argument Organization}} Argument organization is about extracting the core arguments from a set of articles and presenting them in a consolidated manner, making it easier for an end user to quickly understand what authors of the different articles are actually arguing about.
 
\paragraph{\textbf{Community Based Approach for Argument Organization}} A community-based approach to argument organization is about grouping a high volume of arguments into smaller, more focused clusters based on their shared perspectives. By clustering arguments into distinct communities defined by related subtopics or common viewpoints, this method helps break down complex articles into more manageable and coherent segments. It allows for a more nuanced understanding by recognizing that within a larger argument, there are often smaller, interconnected conversations happening among authors with similar views or concerns.
 
\paragraph{\textbf{Understanding Political Stances}}
Left: Arguments that display bias in ways that strongly align with liberal, progressive, or left-wing thought and/or policy agendas\newline
Right: Arguments that display bias in ways that strongly align with conservative, traditional, or right-wing thought and/or policy agendas\newline
Center: Arguments that do not show much political bias or display a balance between left and right perspectives\newline
Here are some arguments pertaining to each stance on the topic of "gun control and gun rights:"\newline
\textbf{Left-leaning Argument}: Stricter gun control laws are necessary to reduce gun violence and prevent mass shootings by implementing comprehensive background checks and limiting access to high-capacity weapons.\newline
\textbf{Right-leaning Argument}: The Second Amendment protects individual gun ownership as a fundamental constitutional right, and law-abiding citizens should be able to own firearms for self-defense and protection. \newline
\textbf{Center-leaning Argument}: Balanced gun policies should respect both constitutional rights and public safety concerns through sensible regulations that address gun violence while preserving responsible gun ownership. \newline\newline
\textbf{Understanding Political Viewpoints}
Based on the aforementioned stances, a community can adopt one of seven viewpoints:
Left-aligned: The majority of the arguments in the community are left-leaning.
Right-aligned: The majority of the arguments in the community are right-leaning.
Center-aligned: The majority of the arguments in the community are center-leaning.
Left and Right-aligned: There is nearly an even distribution between left and right-leaning arguments.
Left and Center-aligned: There is nearly an even distribution between left and center-leaning arguments.
Right and Center-aligned: There is nearly an even distribution between right and center-leaning arguments.
Mixed: There is nearly an even distribution among right, left and center-leaning arguments. [Please, note the difference between Center-aligned and Mixed.] \newline\newline
\textbf{Your Task:} This survey will show you outputs from 3 systems that perform community-based argument organization. Under each system, there can be a maximum of 6 communities and each community may contain up to 10 arguments. Each system processes the same number of arguments but organizes them differently based on its underlying methodology. For example, arguments from Community 1 in System 1 may not be the same as arguments from Community 1 in System 2. Your task is to thoroughly analyze the communities from each system and evaluate them on the metrics described below based on your analysis. \newline\newline
\textbf{Description of Systems}\newline\newline 
\textbf{System 1: Graph View} \newline
It shows arguments as an interactive network of nodes (circles) connected by edges (lines).
Each node contains an argument along with its corresponding article number. Each node will be colored as blue, red or green respectively for "left", "right" and "center" stances. Each edge has a label that shows how two nodes (i.e. arguments) are connected.
Edge labels can be one of the following:
Keyword: The arguments connected by the edge share important words or phrases
Semantic: The arguments connected by the edge share similar meanings  irrespective of keywords presence
Concept: The arguments connected by the edge share the same concept(s) irrespective of keywords presence
Entity: The arguments connected by the edge share the mentions of same organizations, persons or places
Source: The arguments connected by the edge share the same origin, i.e. the same article
You can hover over a node to read the argument and then follow the edges to read the connected arguments. You can also use the zoom controls to adjust the level of detail in each graph.
Each community's description clearly states its political viewpoint and includes a label indicating the perspective of its arguments.\newline\newline 
\textbf{System 2: List View} \newline
It shows arguments in a list format.
Each argument starts with an index number followed by its article number. The index number will be colored as blue, red or green respectively for "left", "right" and "center" stances.
Each community's description clearly states its political viewpoint and includes a label indicating the perspective of its arguments.\newline\newline
\textbf{System 3: List View}\newline
The presentation of this system is same as system 2. All communities are provided as HTML files. To return to the original page, right-click the desired HTML file and select "Open in new tab.".\newline\newline
\textbf{Evaluation Criteria}
\newline
After you finish reviewing the outputs from the three systems, you will be asked to score each system (not individual communities) on a scale of 1 to 5 (5 being the highest, 1 being the lowest) based on the following metrics:
\begin{itemize}
    \item \textbf{Diversity of Viewpoints}: How useful are the communities in a system in bringing out a diversity of viewpoints as well as stances?
    \item \textbf{Informativeness}: How well do the arguments in a community collectively provide comprehensive information about the given topic and its associated viewpoints?
    \item \textbf{Coherence}: How well do the arguments in a community flow and connect with each other to form a meaningful discussion of the topic?
    \item \textbf{Overall Quality}: Overall, how do you rate the system with respect to the way arguments are organized?
\end{itemize}